\documentclass{sig-alternate-2013}

\newfont{\mycrnotice}{ptmr8t at 7pt}
\newfont{\myconfname}{ptmri8t at 7pt}
\let\crnotice\mycrnotice%
\let\confname\myconfname%

\permission{Permission to make digital or hard copies of all or part of this work for personal or classroom use is granted without fee provided that copies are not made or distributed for profit or commercial advantage and that copies bear this notice and the full citation on the first page. Copyrights for components of this work owned by others than ACM must be honored. Abstracting with credit is permitted. To copy otherwise, or republish, to post on servers or to redistribute to lists, requires prior specific permission and/or a fee. Request permissions from Permissions@acm.org.}
\conferenceinfo{KDD'15,}{August 10-13, 2015, Sydney, NSW, Australia.} 
\copyrightetc{\copyright~2015 ACM. ISBN \the\acmcopyr}
\crdata{978-1-4503-3664-2/15/08\ ...\$15.00.\\
DOI: http://dx.doi.org/10.1145/2766XXX.XXXXXXX}

\usepackage{amsmath}
\usepackage{algorithm}
\usepackage{algorithmic}
\usepackage{graphicx}
\usepackage{float}
\usepackage[caption = false]{subfig}
\usepackage{parskip}
\usepackage{url}
\usepackage{dblfloatfix}

\begin{document}
\setlength{\parskip}{0pt}
\setlength{\parindent}{10pt}

\title{Smart Pacing for Effective Online Ad Campaign Optimization}
\numberofauthors{1} 
\author{
       \alignauthor Jian Xu, Kuang-chih Lee, Wentong Li, Hang Qi, and Quan Lu\\
       \affaddr{Yahoo Inc.}\\
       \affaddr{701 First Avenue, Sunnyvale, California 94089}\\
       \email{\{xujian,kclee,wentong,hangqi,qlu\}@yahoo-inc.com}
}

\maketitle
\begin{abstract}
In targeted online advertising, advertisers look for maximizing campaign performance under delivery constraint within budget schedule. Most of the advertisers typically prefer to impose the delivery constraint to spend budget smoothly over the time in order to reach a wider range of audiences and have a sustainable impact. Since lots of impressions are traded through public auctions for online advertising today, the liquidity makes price elasticity and bid landscape between demand and supply change quite dynamically. Therefore, it is challenging to perform smooth pacing control and maximize campaign performance simultaneously. In this paper, we propose a \emph{smart pacing} approach in which the delivery pace of each campaign is learned from both offline and online data to achieve smooth delivery and optimal performance goals. The implementation of the proposed approach in a real DSP system is also presented. Experimental evaluations on both real online ad campaigns and offline simulations show that our approach can effectively improve campaign performance and achieve delivery goals.
\end{abstract}


\section{Introduction}
\begin{figure}[h]
\centering
\captionsetup[subfloat]{%
font=scriptsize,
labelformat=parens,labelsep=space,
listofformat=subparens}
\subfloat[Even pacing]{\includegraphics[width = 1.5in]{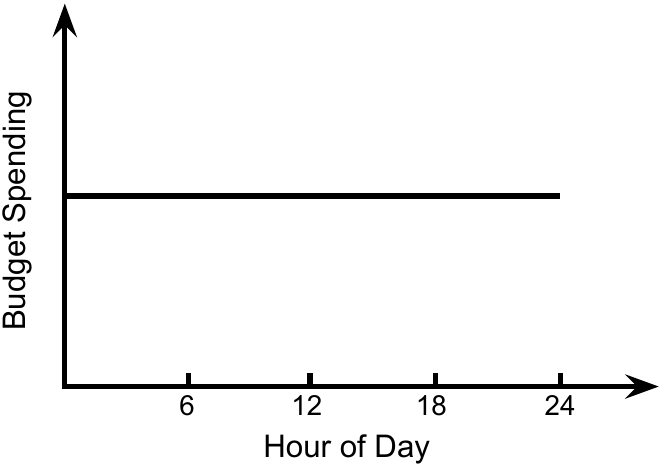}} 
\subfloat[Traffic based pacing]{\includegraphics[width = 1.5in]{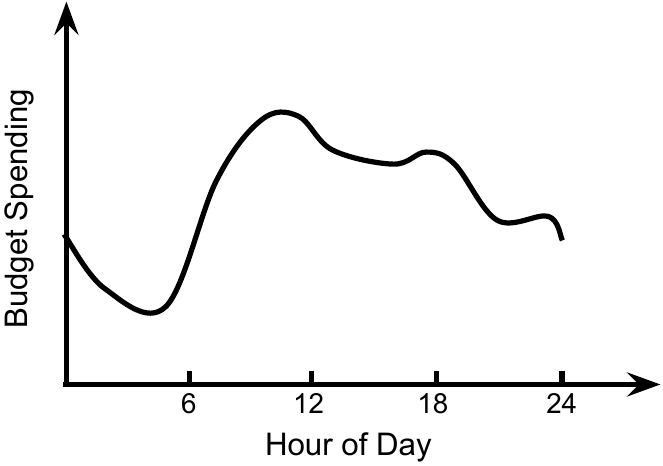}}
\caption{Different budget spending plans.}
\label{fig:pacing_example}
\end{figure}

Online advertising is a multi-billion dollar industry and has been enjoying continued double-digit growth in recent years. The market has witnessed the emergence of search advertising, contextual advertising, guaranteed display advertising, and more recently auction-based advertising exchanges. We focus on the auction-based advertising exchanges, which is a marketplace with the highest liquidity, i.e., each ad impression is traded with a different price through a public auction. In this market, Demand-Side Platforms (DSPs) are key players who act as the agents for a number of different advertisers and manage the overall welfare of the ad campaigns through many direct buying ad-networks or real-time bidding (RTB) ad exchanges in order to acquire different ad impressions. The objectives of an advertiser on a DSP can be summarized as follows:

\begin{itemize}
  \item \emph{Reach the delivery and performance goals:} for \emph{branding} campaigns, the objective is usually to spend out the budget to reach an extensive audience and meanwhile make campaign performance as good as possible; for \emph{performance} campaigns, the objective is usually to meet the performance goal (e.g. eCPC\footnote{\scriptsize Effective cost per click - The cost of a campaign divided by the total number of clicks received. Effective cost per action (eCPA) is defined similarly.} no more than \$2) and meanwhile spend as much budget as possible. Objectives of other campaigns are usually in-between these two extremes.
  \item \emph{Execute the budget spending plan:} advertisers usually expect their ads to be shown smoothly throughout the purchased period in order to reach a wider range of audience, have a sustainable impact, and increase synergy with campaigns on other medias such as TV and magazines. Therefore, advertisers may have their customized budget spending plans. Figure \ref{fig:pacing_example} gives two examples of budget spending plan: even pacing and traffic based pacing.
    \item \emph{Reduce creative serving cost:} apart from the cost to be charged by DSPs, there is also creative serving cost charged by typically 3rd-party creative server providers. This is even more important nowadays that more and more ad campaigns are in the form of video or rich media. The creative serving cost of such type of impressions can be as much as premium inventory cost, so the advertisers will always be willing to reduce this cost and deliver impressions to the right users effectively and efficiently.
\end{itemize}

However, it becomes more and more challenging to achieve all the above objectives simultaneously. In the individual campaign level, each campaign may have its own budget, budget spending plan, targeted audiences, performance goal, billing method, creative serving cost, and so on. In the network level, an increasing number of DSPs compete with each other simultaneously to acquire inventory through public auctions in many ad exchanges, and therefore the price elasticity and bid landscape between demand and supply change dynamically. All those varieties in both campaign and network level make the optimization extremely difficult even for a single campaign. \\
\indent Fortunately, thanks to the rapid growth of emerging internet industries such as mobile apps and user-generated content platforms, the online advertising industry has observed a sharply increasing availability of advertising inventory. A DSP nowadays typically receives tens of billions of ad requests from dozens of Supply-Side Platforms (SSPs) everyday and hence has more flexibility than before to spend budget on a vast amount of advertising opportunities. In this paper, we consider a problem motivated by these trends in the online advertising industry: \emph{can we smartly decide on which inventories the budget should be spent so that all the objectives of an ad campaign are achieved?} We study this problem with a real DSP and explore models and algorithms to effectively serve ad campaigns it manages. Our contributions can be summarized as follows:

\begin{itemize}
  \item We formulate the above multi-objective optimization problem with a comprehensive anatomy of real ad campaigns and propose to solve it through \emph{smart} pacing control.
  \item We develop a control-based method which learns from both online and offline data in order to optimize budget pacing and campaign performance simultaneously.  
  \item We implement the proposed approach in a real DSP and conduct extensive online/offline experimental evaluations. The results show that our approach can effectively improve campaign performance and achieve delivery goals.
\end{itemize}

The rest of the paper is organized as follows: we review the related work in section 2. In section 3, we present the notations and formal problem statement. We describe our models and algorithms in section 4 and 5. Section 6 and 7 focus on experiments and implementation respectively. We conclude our work and discuss future work in section 8.

\begin{figure}
\centering
\captionsetup[subfloat]{%
font=scriptsize,
labelformat=parens,labelsep=space,
listofformat=subparens}
\subfloat[Probabilistic throttling]{\includegraphics[width = 1.5in]{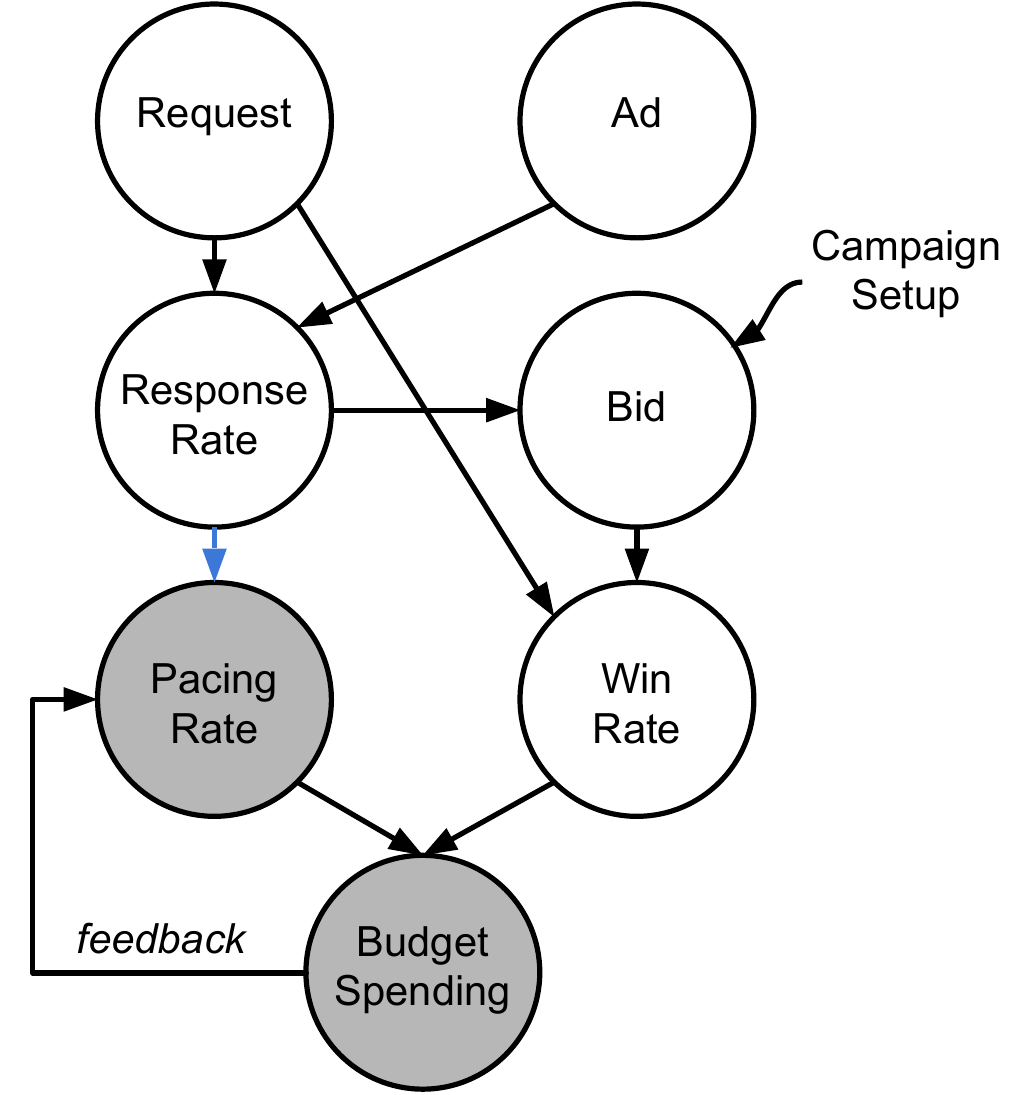}} 
\subfloat[Bid modification]{\includegraphics[width = 1.4in]{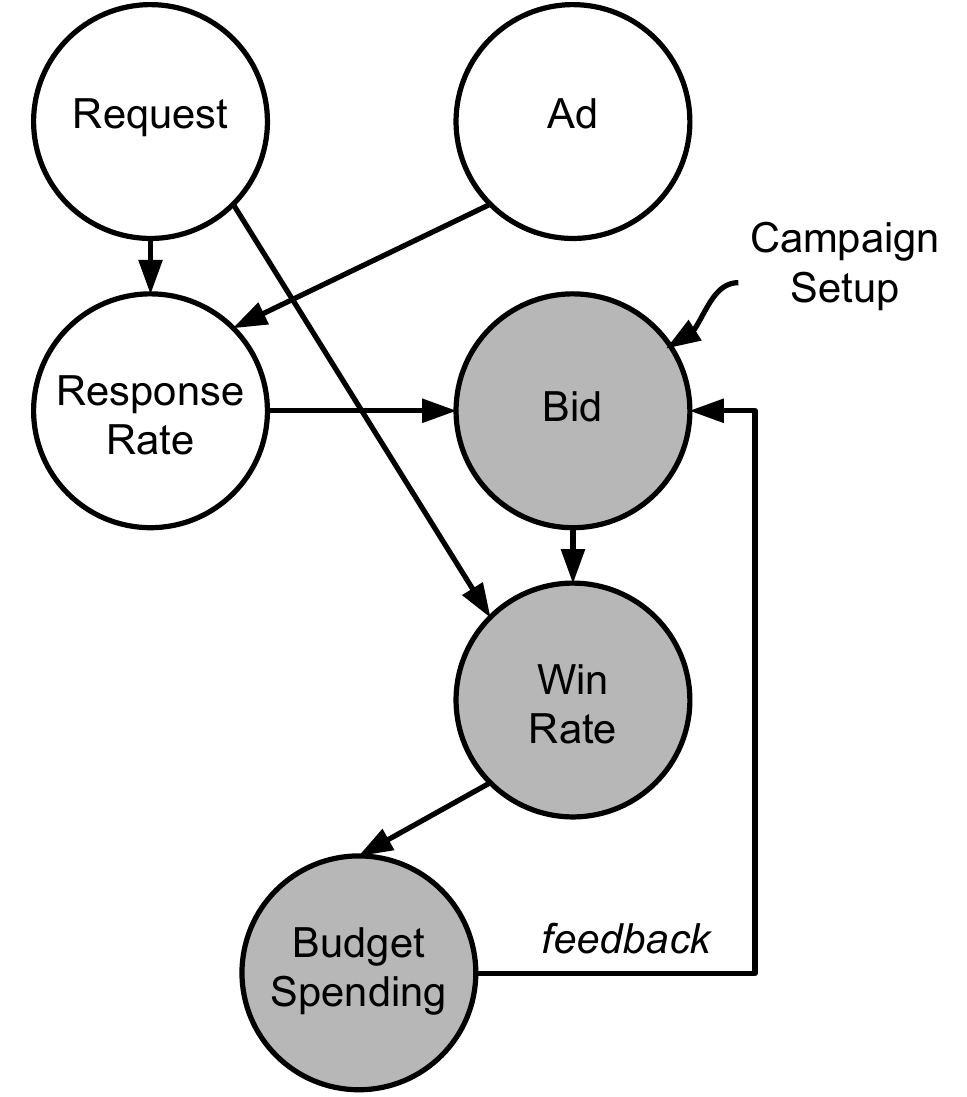}}
\caption{Factor dependency graph in Probabilistic Throttling v.s. Bid Modification. Factors in grey are involved in budget pacing control. Adding dependency between pacing rate and response rate is one of the key ideas of our work.}
\label{fig:throttling}
\end{figure}

\section{Related Work}
\label{sec:related}
Most existing work related to campaign optimization focuses on estimating the click through rate (CTR)/action rate (AR) or bid landscape, which helps to setup the bid price in the impression level. For a comprehensive survey of all those methods, please refer to \cite{OChapelle:CTRPrediction}. However, those approaches do not take into account of the smooth delivery constraint. \\
\indent Another research direction deals with the ad allocation problem. Chen et al. \cite{YChen:AdAllocation} and Bhalgat et al. \cite{VMirrokni:OnlineSmoothDelivery} proposed to perform online ad allocation and solve the optimization problem for all campaigns in the marketplace level. Their focus is to find the best allocation plan to optimize revenue on the publisher side while we focus on maximizing the delivery and performance for each campaign. Bhalgat et al. \cite{ABhalgat:AdAllocation} proposed a different approach to perform ad allocation based on the inventory quality, which shares some common thoughts with our work. We both assume that the inventory is much larger than advertiser demand and therefore both of our solutions focus on how to select high quality inventory. However, their approach focuses on the allocation problem and the smooth delivery constraint does not appear in the formulation. Zhang et al. \cite{WZhang:JointOptimizationBidBudget} proposed to combine budget allocation and bid optimization in a joint manner. We claim that the budget allocation can be complementary to our work in the campaign level. After the budget allocation is done, our approach of budget pacing can take place to further optimize the campaign goals in the impression level.  \\
\indent Mehta et al. \cite{mehta2007adwords}, Abrams et al. \cite{abrams2007optimal}, and Borgs et al. \cite{borgs2007dynamics} suggested using \emph{bid modification} while Agarwal et al. \cite{agarwal2014budget} and Lee et al.\cite{KLee:BudgetPacingTurn} used \emph{probabilistic throttling} to achieve pacing control. In this paper, we use probabilistic throttling for several considerations: 1) Probabilistic throttling directly influences budget spending while bid modification changes the win-rate to control spending, which is not preferred in the RTB environment. First, the bid win-rate curve is usually not smooth so modifying bid can cause significant changes in budget spending. Second, our observation on real serving data is that the bid landscape can be changed dramatically over time. Both issues make pacing control extremely difficult through bid modification. 2) SSPs usually set reserve prices so pacing control may fail if the bid need to be modified to be below the reserve price \cite{agarwal2014budget}. 3) As shown in Figure \ref{fig:throttling}, throttling with a pacing rate decouples pacing control from bid calculation. This is an appealing feature because the pacing control can be developed independently and combined with any bid optimization implementation. In the rest of this paper, we assume the bid is already given by a preceding bid optimization module. 

\section{Problem Formulation}
\label{sec:problem}
We focus on two most prevailing campaign types: 1) branding campaigns that aim at spending out the budget to have an extensive reach of audiences, and 2) performance campaigns that have specific performance goals (e.g. eCPC $\le\$2$). Other types of campaigns typically lie in-between these two extremes. A campaign of either type may have its unique budget spending plan. We first formulate the problem to be resolved and then outline our solution.

\subsection{Preliminaries}
Let $Ad$ be an ad campaign, $B$ be the budget of $Ad$, and $G$ be the performance goal of $Ad$ if there is. A \emph{spending plan} is a sequence of budgets over a number $K$ of time slots, specifying the desired amount of budget to be spent in each time slot. We denote by $\mathbf{B}=(B^{(1)},...,B^{(K)})$ the spending plan of $Ad$, where $B^{(t)}\ge0$ and $\sum_{t=1,...,K}B^{(t)}=B$. Let $Req_i$ be the $i$-th ad request received by a DSP. As we discussed in section \ref{sec:related}, we use probabilistic throttling for budget pacing control in our work. Thus we denote by: \\
\indent $s_i\sim Bern(r_i)$ the variable indicating whether $Ad$ participates the auction for $Req_i$, where $r_i$ is the \emph{point pacing rate} of $Ad$ on $Req_i$. $r_i\in [0,1]$ quantifies the probability that $Ad$ participates the auction for $Req_i$. \\
\indent $w_i$ the variable indicating whether $Ad$ wins $Req_i$ if it participates the auction, which depends on the bid $bid_i$ given by the bid optimization module. \\
\indent $c_i$ the advertiser's cost if $Ad$ is served to ad request $Req_i$. We note that the cost consists of both the inventory cost and the creative serving cost,  \\
\indent $q_i\sim Bern(p_i)$ the variable indicating whether the user performs some desired response (e.g. click) if $Ad$ is served to $Req_i$, where $p_i=Pr(respond|Req_i, Ad)$ is the probability of such response. \\
\indent $C=\sum_i s_i \times w_i \times c_i$ the total cost of ad campaign $Ad$. \\
\indent $P=C/\sum_i s_i \times w_i \times q_i$ the performance (e.g. eCPC if the desired response is click) of ad campaign $Ad$. \\
\indent $\mathbf{C}=(C^{(1)}, \ldots, C^{(K)})$ the \emph{spending pattern} over the $K$ time slots, where $C^{(t)}$ is the cost in the $t$-th time slot, $C^{(t)}\ge0$ and $\sum_{t=1,...,K}C^{(t)}=C$. \\
\indent Given an ad campaign $Ad$ , we define $\Omega$ to be the penalty (error) function that captures how the spending pattern $\mathbf{C}$ is deviated from the spending plan $\mathbf{B}$. A smaller value will indicate a better alignment. As an example, we may define the penalty as follows: 
\begin{equation} \label{eq:penalty}
 \Omega(\mathbf{C}, \mathbf{B}) = \sqrt{\frac{1}{K}\sum_{t=1}^K (C^{(t)} - B^{(t)})^2}
\end{equation}

\subsection{The Problem of Smart Pacing for Online Ad Campaign Optimization}
Advertisers look for spending budget, executing spending plan, and optimizing campaign performance simultaneously. However, there may be multiple \emph{Pareto} optimal solutions to such an abstract multi-objective optimization problem. In real scenarios, advertisers usually prioritize these objectives for different campaigns. For branding campaigns, advertisers typically put budget spending at the top priority, followed by aligning with spending plan while performance is not a serious concern. At serving time (i.e. the ad request time), since we use probabilistic throttling, the only thing that we can have full control is $r_i$. Thus the problem of \emph{\textbf{smart pacing for ad campaigns without specific performance goals}} is defined as determining the values of $r_i$ so that the following measurement is optimized\footnote{\scriptsize Based on our performance definition (i.e. eCPC or eCPA) a smaller value means a better performance.}:
\begin{equation} 
\label{eq:opta}
  \begin{array}{lr}
  \min\limits_{r_i}\ \  P\\ \\
  s.t. \; \; C = B ,\  \Omega(\mathbf{C}, \mathbf{B}) \le \epsilon
  \end{array}
\end{equation}
where $\epsilon$ defines the tolerance level on deviating from the spending plan. On the contrary, for performance campaigns that have specific performance goals, achieving the performance goal is the top priority. Sticking to the spending plan is usually the least important consideration. We define the problem of \emph{\textbf{smart pacing for ad campaigns with specific performance goals}} as determining the values of $r_i$ so that the following measurement is optimized:
\begin{equation} 
\label{eq:optb}
  \begin{array}{lr}
  \min\limits_{r_i}\ \  \Omega(\mathbf{C}, \mathbf{B})\\ \\
  s.t.\;\; P \le G, \ B-C\le\varepsilon
  \end{array}
\end{equation}
where $\varepsilon$ defines the tolerance level for not spending out all the budget. Given the dynamics of the marketplace, even both the single-objective optimization problems are extremely difficult to resolve. Existing methods that are widely used in the industry deal only with capturing either the performance goal or the budget spending goal. One example for achieving performance goal is always bidding the retargeting beacon triggered ad requests. Unfortunately, there is no guarantee to avoid overspending or underspending. Another example for smooth pacing control is introducing a global pacing rate so that all ad requests have the same probability to be bid by a campaign. However, none of these existing approaches can solve the smart pacing problem we formulate here. To tackle this problem, we examine the prevailing campaign setups and make some key observations that motivated our solution: 

\begin{itemize}
  \item \emph{CPM campaigns}: advertisers are charged a fixed amount of money for each impression. For branding advertisers, the campaign optimization is as defined in Equation \ref{eq:opta}. As long as budget can be spent and spending pattern is aligned as the plan, high responding ad requests should have a higher point pacing rate than low responding ones so that the performance can be optimized. For performance advertisers (i.e. with eCPC, eCPA goal), the campaign optimization is as defined in Equation \ref{eq:optb}. Apparently, high responding ad requests should have higher point pacing rate to achieve the performance goal. 
  \item \emph{CPC/CPA campaigns}: advertisers are charged based on the sheer number of clicks/actions. There is implicit performance goal to guarantee that DSP does not lose money when bidding on behalf of the advertisers. So it falls in the category of optimization defined in Equation \ref{eq:optb}. Granting high responding ad requests high point pacing rates will be more effective from both the advertisers' and DSPs' perspectives: advertisers pay less on creative serving cost while DSP can save more ad opportunities to serve other campaigns.
  \item \emph{Dynamic CPM campaigns}: DSP charges a dynamic amount of money for each impression instead of a fixed amount. These campaigns usually have specific performance goals so the optimization problem falls in Equation \ref{eq:optb}. Similar with CPC/CPA campaigns, high responding ad requests are more preferred in order to reduce creative serving cost and save ad opportunities.
\end{itemize}

\subsection{Solution Summary}
Motivated by these observations, we develop novel heuristics to solve the smart pacing problem. The heuristics try to find a feasible solution that satisfies all constraints as defined in Equation \ref{eq:opta} or \ref{eq:optb}, and then further optimize the objectives through feedback control. We first learn from offline serving logs to build a response prediction model to estimate $p_i = Pr(respond|Req_i, Ad)$, which helps distinguish high responding ad requests from low responding ones. Second, we reduce the solution space by grouping similarly responding ad requests together and the requests in the same group share the same \emph{group pacing rate}. Groups with high responding rates will enjoy high pacing rates (refer to the blue arrow in Figure \ref{fig:throttling}(a)). Third, we develop a novel control-based method to learn from online feedback data and dynamically adjust the group pacing rates to approximate the optimal the solution. Without loss of generality, we assume campaign setup is CPM billing with or without an eCPC goal. Our approach can be applied to other billing methods and performance types as well as other grouping strategies such as grouping based on $p_i/c_i$ (the expected response per cost). 

\section{Response Prediction}
\label{sec:performance}
Our solution depends on an accurate response prediction model to estimate $p_i$. There are plenty of work in the literature addressing this problem as we have reviewed in section \ref{sec:related}. Here we briefly describe how we perform this estimation. We use the methodology introduced in \cite{Agarwal:EstimatingRates2,KLee:EstimateCVRTurn} and make some improvements on top of that. In this method, we first leverage the hierarchy structures in the data to collect response feedback features at different granularities. For example, at ad side, starting from the root and continuing layer after layer are \emph{advertiser category, advertiser, campaign}, and finally \emph{ad}. Historical response rates at different levels in the hierarchy structures are used as features to derive a machine learning model (e.g. LR, GBDT, etc) to give a raw estimation of $p_i$, say $\hat{p}_i$. Then we utilize attributes such as user's age, gender to build a shallow tree. Each leaf node of the tree identifies a disjoint set of ad requests which could hardly be further split into subsets with significantly different average response rates. Finally, we calibrate $\hat{p}_i$ within the leaf node $Req_i$ is classified using a piecewise linear regression to estimate the final $p_i$. This scheme results in a fairly accurate response prediction.
\section{A Control-Based Solution}
\label{sec:control}
As we have discussed in section \ref{sec:problem}, it is extremely difficult to reach the exact optimal solution to the problems defined in equations \ref{eq:opta} and \ref{eq:optb} in an online environment. We explore heuristics to reduce the solution space of the original problems. More specifically, with the response prediction model described in section \ref{sec:performance}, similarly responding ad requests are grouped together and they share the same \emph{group pacing rate}. Different groups will have different group pacing rates to reflect our preferences on high responding ad request groups. The original problem of solving the point pacing rate of each $r_i$ is reduced to solving a set of group pacing rates. We employ a control-based method to tune the group pacing rates so that online feedback data can be leveraged immediately for campaign optimization. In other words, the group pacing rates are dynamically adjusted throughout the campaign life time. For simplicity, \emph{pacing rate} and \emph{group pacing rate} are interchangeable in the rest of this paper, and we denote by $r_l$ the group pacing rate of the $l$-th group.  
\subsection{A Layered Presentation}
\label{sec:layered}
For each ad campaign, we maintain a layered data structure in which each layer corresponds to an ad request group. We keep the following information of each ad request group in the layered structure: average $response\ rate$ (usually in the form of CTR, AR, etc) derived from the response prediction model; $priority$ of the ad request group; $pacing\ rate$ i.e. the probability to bid an ad request in the ad request group; and the campaign's $spending$ on the ad request group in the latest time slot. The principles here are: 1) layers correspond to high responding ad request groups should enjoy high priorities, and 2) the pacing rate of a high priority layer should not be smaller than that of a low priority layer. \\
\indent For each campaign, when the DSP receives an eligible ad request, it first decides which ad request group the ad request falls in and refers to the corresponding layer to acquire the pacing rate. The DSP then bids the ad request on behalf of the campaign with a probability that equals to the retrieved pacing rate at the price given by a preceding bid optimization module.

\subsection{Online Pacing Rate Adjustment}
We employ a control-based method to adjust the pacing rate of each layer based on real-time feedbacks. Suppose we have $L$ layers, the response rate estimation by response prediction model for each layer is $\mathbf{p}=(p_1,\ldots,p_L)$, and hence if the desired response is click, the estimated eCPC of each layer is $\mathbf{e}=(e_1,\ldots,e_L)$ where $e_i=\frac{CPM}{1000\times p_i}$. Let the pacing rate of each layer in the $(t-1)$-th time slot be $\mathbf{r}^{(t-1)}=(r_1^{(t-1)},\ldots,r_L^{(t-1)})$, and the spending of each layer be $\mathbf{c}^{(t-1)}=(c_1^{(t-1)},$ $\ldots,c_L^{(t-1)})$, the control-based method will derive $\mathbf{r}^{(t)}=(r_1^{(t)},$ $\ldots,r_L^{(t)})$ for the coming $t$-th time slot based on campaign objectives.

\begin{algorithm}[t]
\caption{AdjustWithoutPerformanceGoal}
\label{alg:A}
\begin{algorithmic}[1]
\REQUIRE $\mathbf{c}^{(t-1)}$, $\mathbf{r}^{(t-1)}$, $R$
\ENSURE $\mathbf{r}^{(t)}$
\IF {$R==0$}
    \STATE return $\mathbf{r}^{(t)}=\mathbf{r}^{(t-1)}$
\ELSIF {$R > 0$}
    \FOR {each layer $l$ in $(L,\ldots,l^{'})$}
        \STATE $r_l^{(t)} = min(1.0, r_l^{(t-1)} \times \frac{c_l^{(t-1)} + R}{c_l^{(t-1)}})$
        \STATE $R = R - c_l^{(t-1)}\times\frac{r_l^{(t)}-r_l^{(t-1)}}{r_l^{(t-1)}}$
    \ENDFOR
    \STATE $r_{l^{'}-1}^{(t)} = trial\ rate$ \textbf{if}  $l^{'} \ne 1$ \textbf{and}  $r_{l^{'}}^{(t)} > trial\ rate$
\ELSE
    \FOR {each layer $l$ in $(l^{'},\ldots,L)$}
        \STATE $r_l^{(t)} = max(0.0, r_l^{(t-1)} \times \frac{c_l^{(t-1)} + R}{c_l^{(t-1)}})$
        \STATE $R = R - c_l^{(t-1)}\times\frac{r_l^{(t)}-r_l^{(t-1)}}{r_l^{(t-1)}}$
        \IF {$R \ge 0$}
            \STATE $r_{l-1}^{(t)} = trial\ rate$ \textbf{if} $l \ne 1$ \textbf{and}  $r_l^{(t)} > trial\ rate$
            \STATE \textbf{break}
        \ENDIF
    \ENDFOR
\ENDIF
\RETURN $\mathbf{r}^{(t)}=(r_1^{(t)},\ldots,r_L^{(t)})$
\end{algorithmic}
\end{algorithm}

\subsubsection{Campaigns without Performance Goals}
\begin{figure}[t]
  \centering
  \includegraphics[width=2.8in]{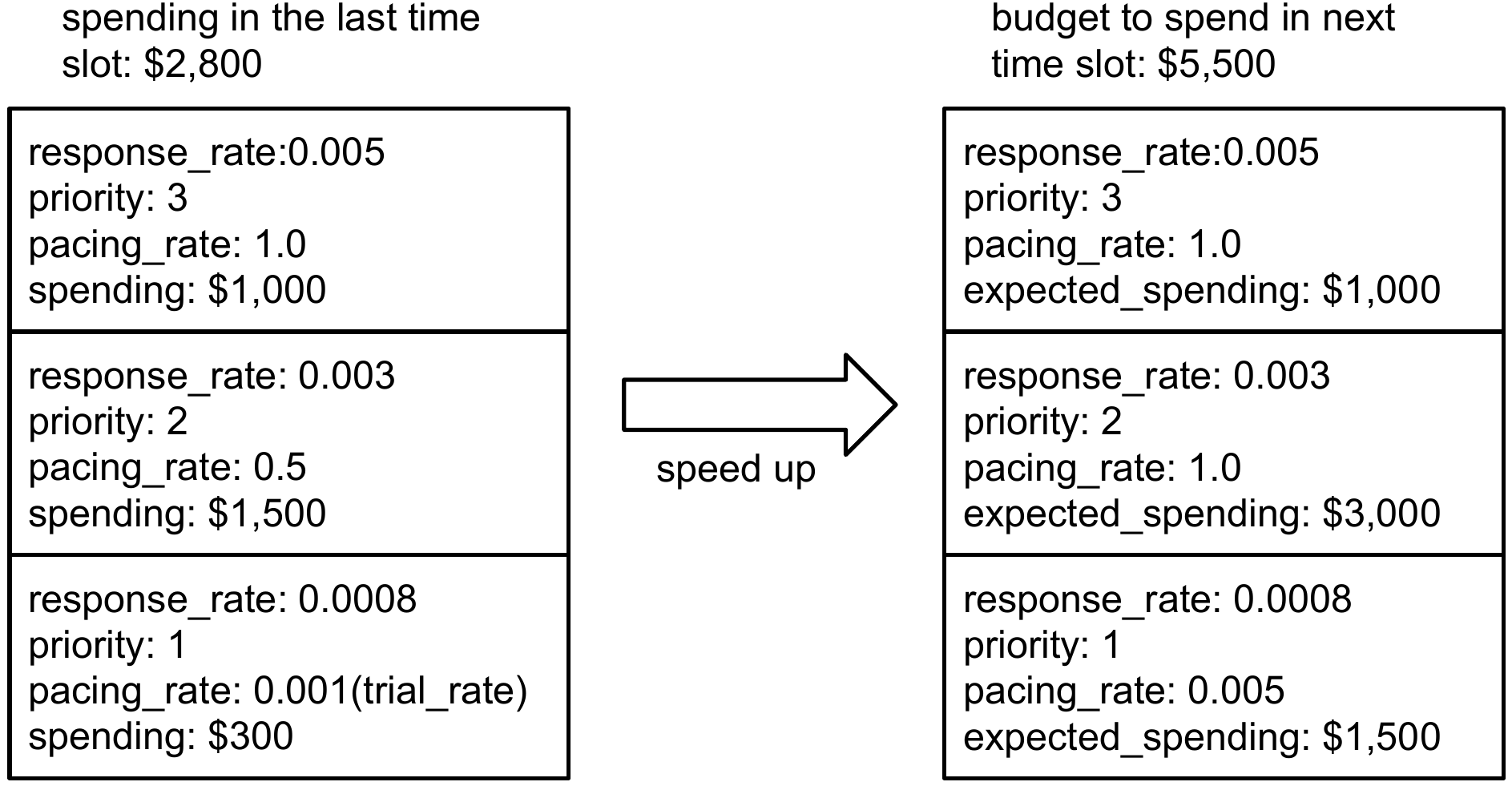}
  \caption{An example to speed up budget spending}
   \label{fig:speedup}
\end{figure}

\begin{figure}[t]
  \centering
  \includegraphics[width=2.8in]{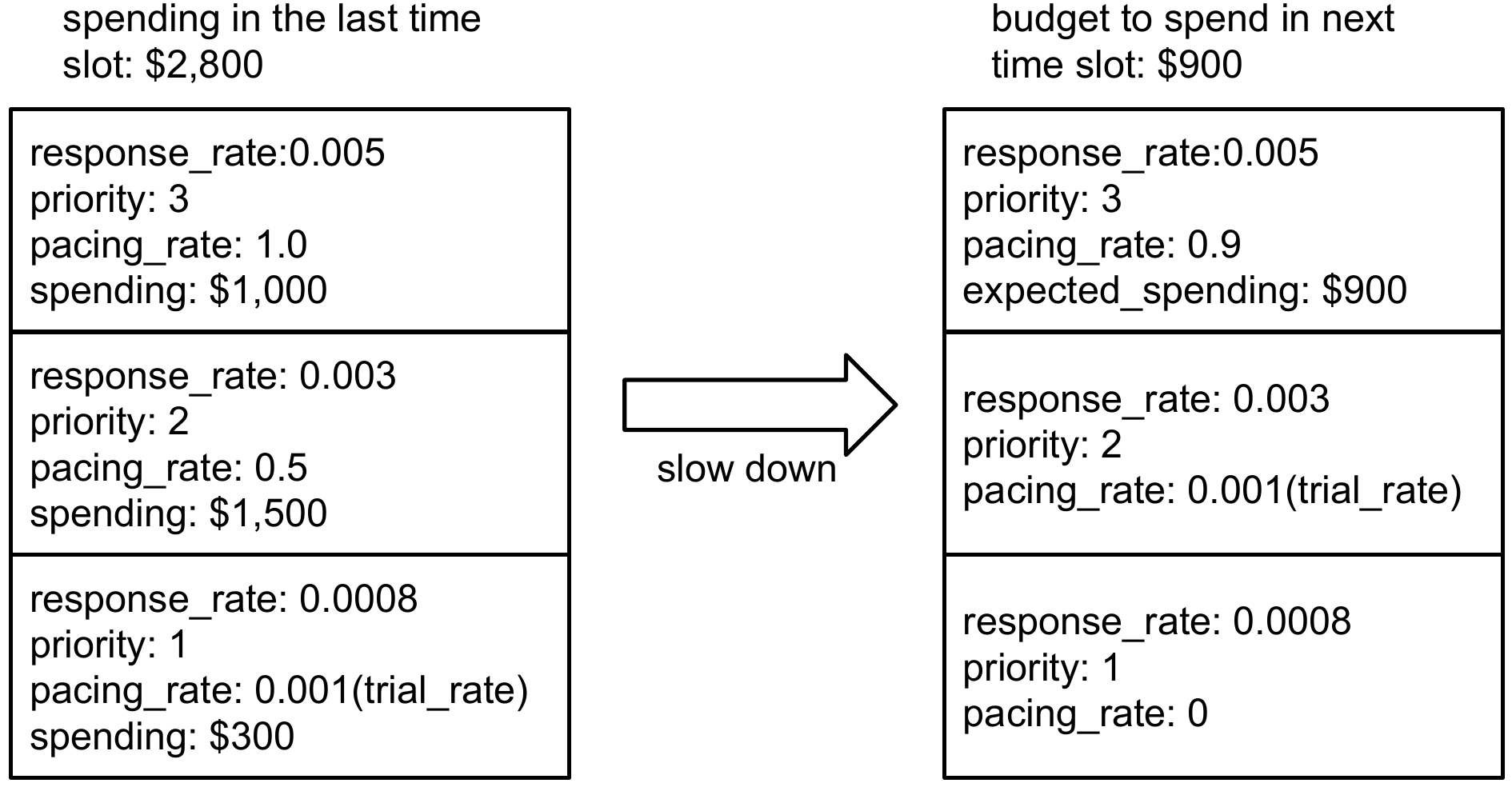}
  \caption{An example to slowdown budget spending}
   \label{fig:slowdown}
\end{figure}

We first describe the adjustment algorithm for ad campaigns without specific performance goals. Recall that for such campaign type, the primary goal is to spend out the budget and align with the budget spending plan. Thus at the end of each time slot, the algorithm needs to decide the amount of budget to be spent in the next time slot and adjust the layered pacing rates to spend exact that amount. \\
\indent The budget to be spent in the next time slot is determined based on the current budget spending status. Given an ad campaign, suppose its total budget is $B$, budget spending plan is $\mathbf{B}=(B^{(1)},...,B^{(K)})$, and after running for $m$ time slots, the remaining budget becomes $B_m$. We need to decide the desired spending in each of the remaining time slots, denoted as $\widehat{C}^{(m+1)} \ldots \widehat{C}^{(K)}$ so that the total budget can be spent out and the penalty is minimized.
\begin{equation}
\label{desired_spending_abstract}
  \begin{array}{lr}
  \operatorname*{arg\,min}\limits_{\widehat{C}^{(m+1)},\ldots,\widehat{C}^{(K)}} \Omega \\ \\
  s.t. \; \; \sum\limits_{t=m+1}^K\widehat{C}^{(t)} = B_m
  \end{array}
\end{equation}
\noindent in which if we adopt the definition of $\Omega$ as in equation \ref{eq:penalty}, we have the following optimal solution:
\begin{equation}
\label{desired_spending_solution}
  \widehat{C}^{(t)}=B^{(t)} + \frac{B_m - \sum_{t=m+1}^K {B^{(t)}}}{K-m}
\end{equation}
\noindent where $t=m+1,\ldots,K$. We omit the details of how $\widehat{C}^{(t)}$ is derived because of page limit. In an online environment, suppose the actual spending in the latest time slot is $C^{(t-1)}$, we define $R = \widehat{C}^{(t)} - C^{(t-1)}$ to be the $residual$ which can help us to make adjustment decisions. \\
\indent Algorithm \ref{alg:A} gives the details of how the adjustment is done. Suppose index $L$ represents the highest priority, index $1$ refers to the lowest priority, and let $l^{'}$ be the last layer with non-zero pacing rate (the principles in section \ref{sec:layered} guarantee the existence of $l^{'}$). If $R$ equals $0$, no adjustment is needed. If $R > 0$, which means delivery should speed up, pacing rates are adjusted in a top-down fashion. Starting from layer $L$, the pacing rate of each layer is increased one-by-one until layer $l^{'}$. Line 5 calculates the desired pacing rate of the current layer in order to offset $R$. We give layer $l^{'}-1$ a \emph{trial rate} to prepare for future speedups if layer $l^{'} \ne 1$ and its updated pacing rate $r_{l^{'}}^{(t)} > trial\ rate$. Figure \ref{fig:speedup} gives an example of how the speedup adjustment is done. If $R < 0$, which means delivery should slow-down, pacing rate of each layer is decreased in a bottom-up fashion until $R$ is offset. Line 11 derives the desired pacing rate of the current layer to offset $R$. Suppose $l$ is the last layer adjusted, $l\ne 1$ and its new pacing rate $r_l^{(t)} > trial\ rate$, we give layer $l-1$ the trial rate to prepare for future speedups. Figure \ref{fig:slowdown} is an example how delivery is slowed down. \\
\indent We note that this greedy strategy tries to approach the optimal solution to Equation \ref{eq:opta} in an online environment. Within each time slot, it strives to invest on inventories with the best performance under the total budget and spending plan constraints. 

\begin{algorithm}[t]
\caption{AdjustWithPerformanceGoal}
\label{alg:B}
\begin{algorithmic}[1]
\REQUIRE $\mathbf{c}^{(t-1)}$, $\mathbf{r}^{(t-1)}$, $R$, $\mathbf{e}$, $goal$
\ENSURE $\mathbf{r}^{(t)}$
\STATE $\mathbf{r}^{(t)}$=AdjustWithoutPerformanceGoal($\mathbf{c}^{(t-1)}$, $\mathbf{r}^{(t-1)}$, $R$)
\IF {ExpPerf($\mathbf{c}^{(t-1)},\mathbf{r}^{(t-1)},\mathbf{r}^{(t)},\mathbf{e},1)>goal$}
    \FOR {each layer $l$ in $(1,\ldots,L)$}
        \IF {ExpPerf($\mathbf{c}^{(t-1)}, \mathbf{r}^{(t-1)},\mathbf{r}^{(t)},\mathbf{e},l+1)>goal$}
            \STATE $r_l^{(t)}=0.0$
        \ELSE
             \STATE $r_l^{(t)}=r_l^{(t-1)}\times\frac{\sum_{i=l+1,\ldots,L} c_i^{(t-1)}\times(\frac{goal}{e_i}-1)}{c_l^{(t-1)}\times(1-\frac{goal}{e_l})}$
             \IF {$l \ne 1$}
                 \STATE $r_{l-1}^{(t)} = trial\ rate$
             \ENDIF
             \STATE \textbf{break}
        \ENDIF
    \ENDFOR
\ENDIF
\RETURN $\mathbf{r}^{(t)}=(r_1^{(t)},\ldots,r_L^{(t)})$
\end{algorithmic}
\end{algorithm} 

\subsubsection{Campaigns with Performance Goals}
For campaigns with specific performance goals (e.g. eCPC $\le \$2$), pacing rate adjustment is a bit complicated. It is difficult to foresee the ad request traffic in all the future time slots and the response rate distribution can be time-varying. Therefore, given budget spending objectives, exploiting all the ad requests in current time slot that meet performance goal may not be an optimal solution to Equation \ref{eq:optb}. Algorithm \ref{alg:B} describes how the adjustment is done for such kind of campaigns. We adopt the heuristic that a further adjustment based on performance goal is appended to Algorithm \ref{alg:A}. If the expected performance after Algorithm \ref{alg:A} does not meet the performance goal, the pacing rates are reduced one-by-one from the low priority layers until the expected performance meets the goal. Line 7 derives the desired pacing rate of current layer to make the overall expected eCPC meet the goal. Function ExpPerf($\mathbf{c}^{(t-1)},\mathbf{r}^{(t-1)},\mathbf{r}^{(t)},\mathbf{e},i$) in Line 2 and 4 estimates the expected joint eCPC of layers $i,\ldots,L$ if pacing rates are adjusted from $\mathbf{r}^{(t-1)}$ to $\mathbf{r}^{(t)}$, where $e_j$ is the eCPC of layer $j$. 
\begin{equation}
\label{exp_perf}
  \text{ExpPerf(}\mathbf{c}^{(t-1)},\mathbf{r}^{(t-1)},\mathbf{r}^{(t)},\mathbf{e},i\text{)}=\frac{\sum\limits_{j=i}^L\frac{c_j^{(t-1)}\times r_j^{(t)}}{r_j^{(t-1)}}}{\sum\limits_{j=i}^L\frac{c_j^{(t-1)}\times r_j^{(t)}}{r_j^{(t-1)}\times e_j}}
\end{equation}

\subsection{Number of Layers, Initial and Trial Rates}
It is important to set proper number of layers, initial and trial pacing rates. For a new ad campaign without any delivery data, we identify the most similar existing ad campaigns in our DSP and estimate a proper global pacing rate $r_G$ at which we expect the new campaign can spend out its budget. Then the number of layers is set as $L=\lceil\frac{1}{r_G}\rceil$. We note that a moderate number of layers is more desirable than an excessive one for two reasons: 1) the delivery statistics of each layer is not significant if there are too many layers; 2) from the system perspective, excessive number of layers may use up bandwidth and/or memory. \\
\indent Once the number of layers is determined, we run the campaign at the global pacing rate $r_G$ in the first time slot. We call this step an \emph{initialization phase} in which the delivery data can be collected. We group equal amount of impressions into the desired number of layers based on their predicted response rate to identify the layer boundaries. In the next time slot, the pacing rate of each layer is reassigned based on planned budget in the next time slot and high responding layers will have rates of $1.0$ while low responding ones will have rates of $0.0$. \\
\indent In the adjustment algorithms, the direct successive layer next to the layer with non-zero pacing rate is assigned a trial pacing rate. The purpose is to collect delivery data in this layer and prepare for future speedups. This trial rate is supposed to be quite low. We derive such a rate by reserving a certain portion $\lambda$ (e.g. $\lambda=1\%$) of budget to be spent in the next time slot. Let the trial layer be the $l$-th layer and the budget for next time slot is $\widehat{C}^{(t)}$, recall that we have historical spending and pacing rate of this layer from at least one time slot (the initialization phase), the trial pacing rate is derived as $trial\ rate=r_l^{(*)}\times \frac{\lambda \times \widehat{C}^{(t)} }{c_l^{(*)}}$, where $c_l^{(*)}$ and $r_l^{(*)}$ are the historical spending and pacing rate of the $l$-th layer. \\
\indent As a quick summary, we employ a layered presentation of all the ad requests based on their predicted response rate and execute budget pacing control in the layer level to achieve delivery and performance goals. Both the spending in the current time slot and the remaining budget are considered to calculate the layered pacing rates in the next time slot. We also tried the alternative to control a threshold so that only ad requests with predicted response rate above the threshold were bid. The outcome of such alternative, however, was not satisfactory. The main reason is that ad requests are usually not smoothly distributed over response rate, and therefore it is difficult to realize smooth control with a single threshold.

\begin{figure*}[!htb]
\centering
\captionsetup[subfloat]{%
font=scriptsize,
labelformat=parens,labelsep=space,
listofformat=subparens}
\subfloat[Campaign 1: budget:\$3K, CPM:\$0.5, even \newline pacing.]{\includegraphics[width = 2.35in]{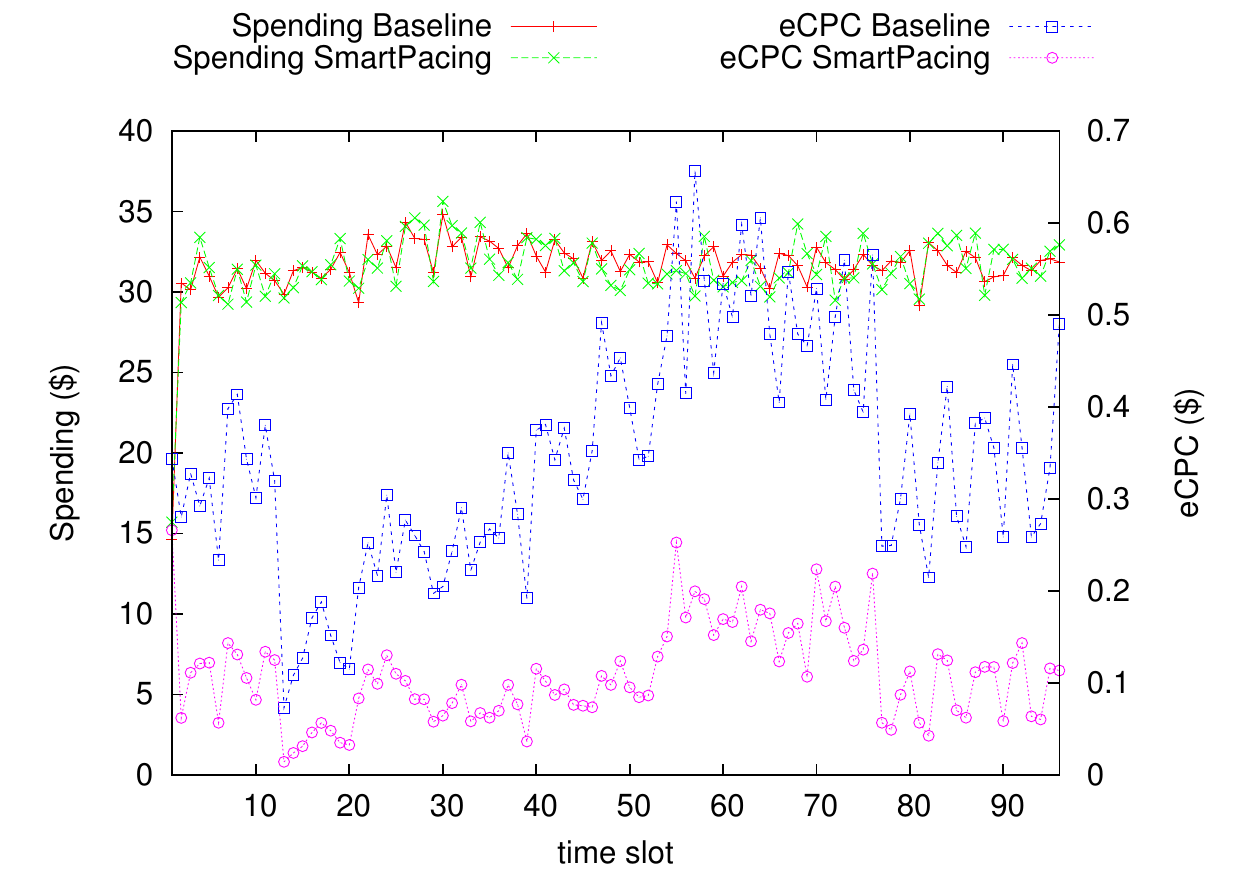}} 
\subfloat[Campaign 2: budget:\$3.8K, CPM:\$2.9, even \newline pacing, search retargeting, eCPC goal:\$2.5.]{\includegraphics[width = 2.35in]{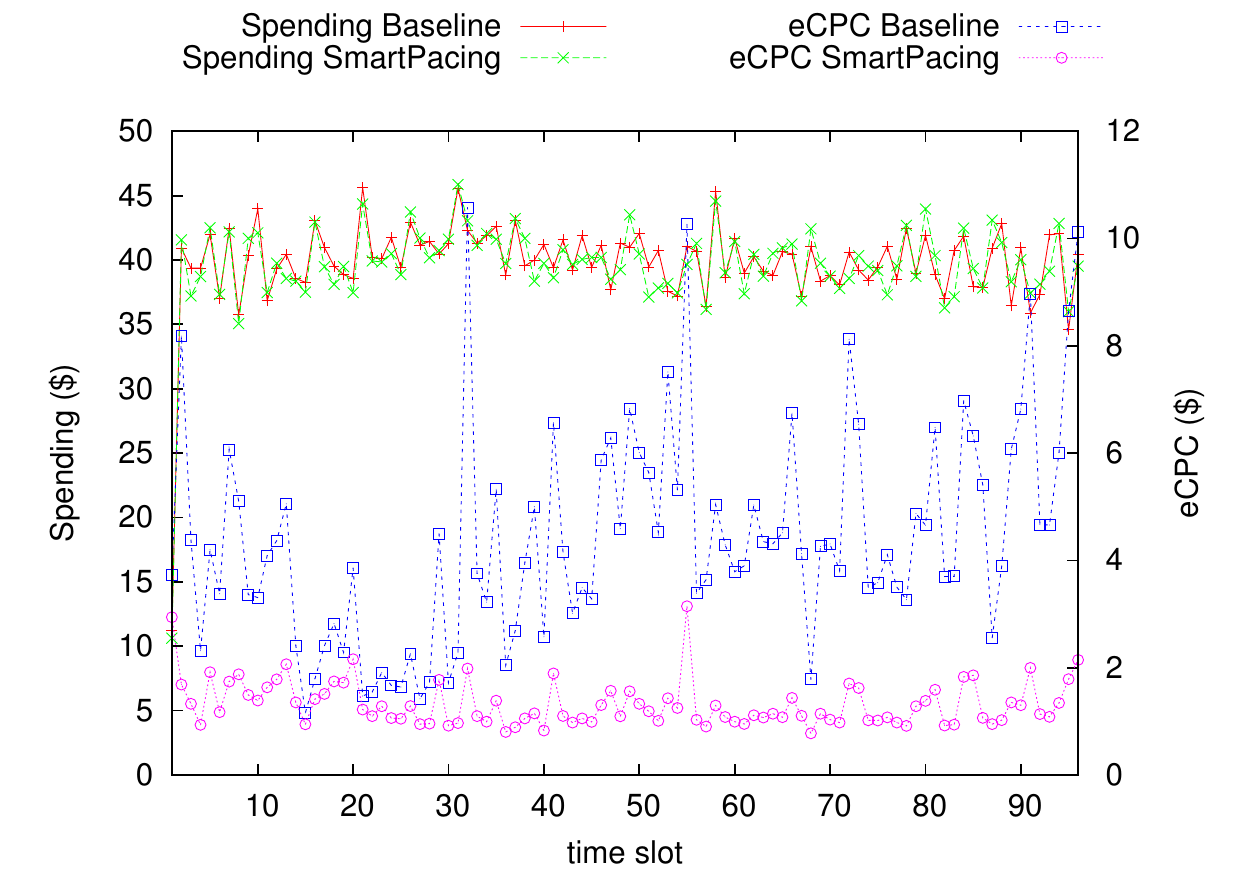}} 
\subfloat[Campaign 3: budget:\$1K, CPM:\$1.5, even pacing, demographic targeting.]{\includegraphics[width = 2.35in]{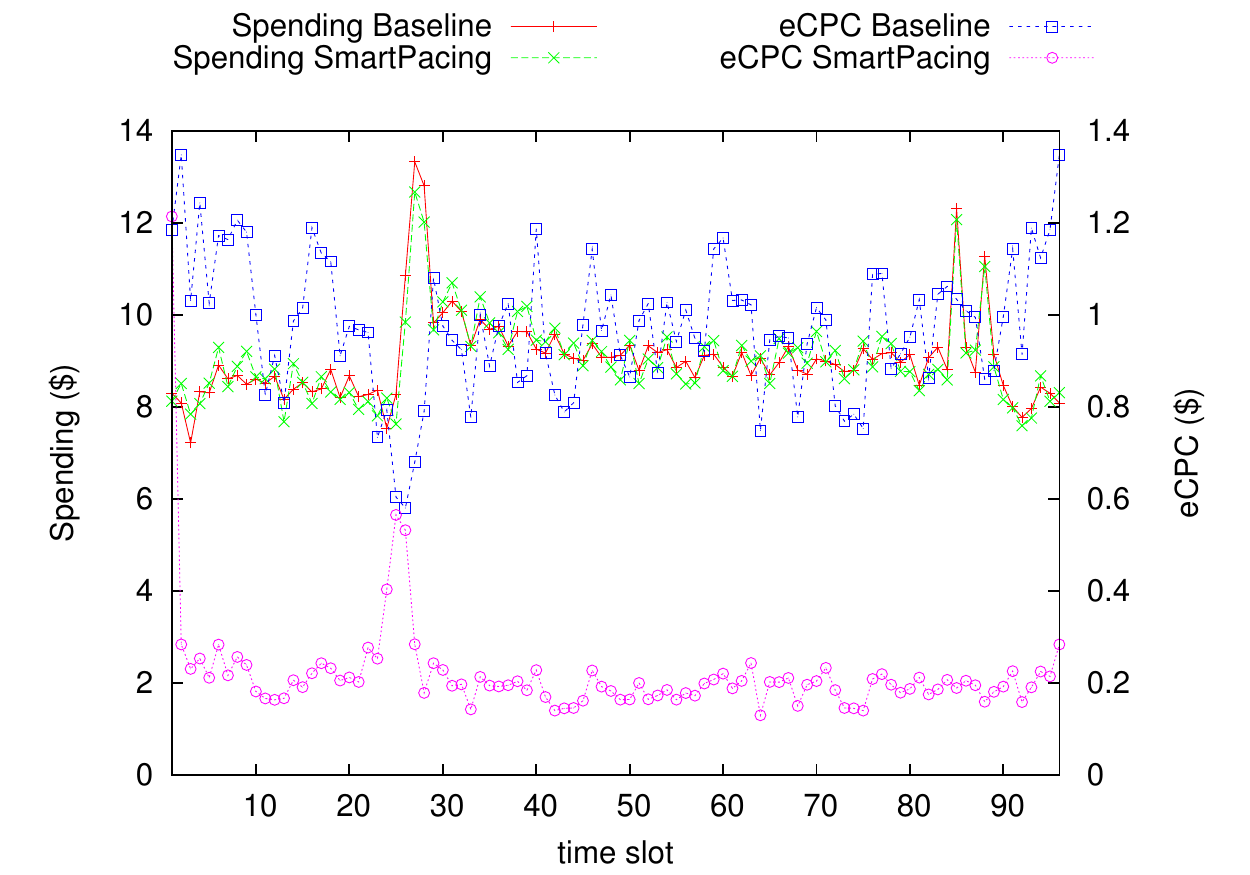}}
\caption{Online A/B test result of 3 real campaigns running on our DSP system (\# layers: 8).}
\label{exp:ab}
\end{figure*}

\begin{table*}[t]
\centering
\resizebox{\textwidth}{!}{%
\begin{tabular}{|c|c|c|c|c|c|c|c|c|c|c|c|c|c|c|}
\hline
\multicolumn{4}{|c|}{\textbf{Campaign Setup}} & \multicolumn{5}{c|}{\textbf{Baseline (1st week)}} & \multicolumn{5}{c|}{\textbf{Smart Pacing (2nd week)}} &  \\ \hline
ID & Budget & CPM & eCPC goal & Spending & Omega & AvgErr & eCPC & AvgPR & Spending & Omega & AvgErr & eCPC & AvgPR & \textbf{eCPC reduction} \\ \hline
4 & \$35K & \$3.10 & \$7.00 & \$35.2K & 20.43 & 5.6\% & \$10.20 & 0.0024 & \$35.3K & 20.72 & 5.7\% & \$5.93 & 0.0026 & \textbf{-41.9\%} \\ \hline
5 & \$20K & \$3.00 & No goal & \$20.4K & 12.21 & 5.9\% & \$14.78 & 0.019 & \$20.2K & 12.32 & 5.9\% & \$7.51 & 0.017 & \textbf{-49.19\%} \\ \hline
6 & \$2K & \$3.35 & No goal & \$1.97K & 2.89 & 13.9\% & \$10.56 & 0.017 & \$2.01K & 2.79 & 13.4\% & \$8.61 & 0.015 & \textbf{-18.47\%} \\ \hline
7 & \$7.5K & \$3.00 & No goal & \$7.37K & 9.08 & 11.6\% & \$9.90 & 0.59 & \$7.58K & 9.19 & 11.8\% & \$9.02 & 0.65 & \textbf{-8.89\%} \\ \hline
\end{tabular}
}
\caption{Online over-time test result of 4 real campaigns (\# layers: 3).}
\label{exp:overtime}
\end{table*}

\section{Experimental Evaluations}
We conduct extensive experiments on both real ad campaigns and offline simulations to evaluate the effectiveness of our approach. Without further specification, the following setting is used throughout the experiments: the baseline approach is our approach with only one layer (i.e. using a global pacing rate), the timespan and time slot interval are 24 hours and 15 minutes respectively, the spending plan is even pacing, and the initial pacing rate and trial budget fraction in our approach are $r_G=0.01$ and $\lambda=1\%$ respectively. We look at all the three aspects discussed throughout this paper: 1) performance, 2) budget spending, and 3) spending pattern. Since the value of the penalty $\Omega$ as defined in Equation \ref{eq:penalty} is hard to interpret, we transform it into the following metric: $AvgErr=(\frac{B}{K})^{-1} \times \Omega$, which quantifies the relative deviation of the actual spending from the average planned spending per time slot, where $K$ is the number of time slots and $B$ is the total budget.

\subsection{Results from Real Campaigns}
We pick up from our DSP system 3 campaigns for online A/B test and another 4 campaigns for online over-time test. All these 7 campaigns are CPM campaigns and two of them have specific eCPC goals. \\
\indent In the A/B test, we deploy 8 layers of group pacing rates in our approach. Campaign 1, 3 are without performance goals while campaign 2 has an eCPC goal of \$2.5.  Figure \ref{exp:ab} shows the test result for these campaigns. Our approach is surprisingly effective in boosting the performance. Compared to the baseline, the eCPC reductions are $-72\%$, $-67\%$, and $-79\%$ for the three campaigns respectively. We note that for Campaign 2, the baseline solution fails to meet the eCPC goal. On the other hand, the total spendings and spending patterns of our approach are as good as the baseline. The $AvgErr$ comparisons of the baseline and our approach are $6.4\%:6.8\%$, $9.1\%:9.2\%$, and $10.2\%:9.8\%$ for the three campaigns respectively. Campaign 3 has experienced spending fluctuations and we found there was another competing campaign went offline at time slot 25 so the win-rate of Campaign 3 surged. Our algorithm can recover very soon from the environment changes and continue to deliver smoothly. \\
\indent The result of over-time test is summarized in Table \ref{exp:overtime}. In this test, we run 4 campaigns for 2 weeks with baseline solution in the first week and our approach with 3 layers in the second week. Among the 4 campaigns, only Campaign 4 has a specific eCPC goal of \$7. Our approach successfully reduces the eCPC of all the 4 campaigns. The most significant reduction almost comes to $-50\%$. Our approach does not show significant eCPC reduction on Campaign 7 because its average pacing rate $AvgPR$ is already around 0.6 and there is not much room to improve its performance. We note that the baseline fails to achieve the eCPC goal of Campaign 4. In real applications, compromising the performance goal may permanently lose the advertiser, which should always be avoided. From the delivery perspective, both the baseline and our approach manage to spend out the total budget smoothly with less than 13.9\% deviation from the spending plan. 
\begin{figure*}[!htb]
\centering
\captionsetup[subfloat]{%
font=scriptsize,
labelformat=parens,labelsep=space,
listofformat=subparens}
\subfloat[Ad request distribution over time slots]{\includegraphics[width = 3.5in]{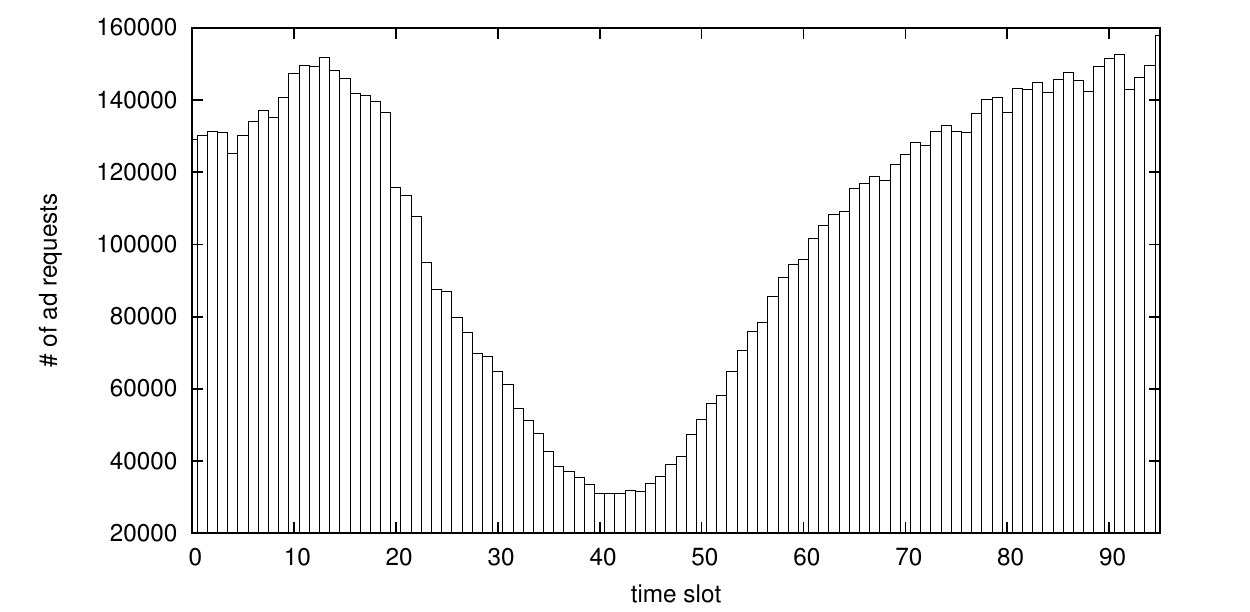}} 
\subfloat[Ad request and win-rate distribution over predicted CTR]{\includegraphics[width = 3.5in]{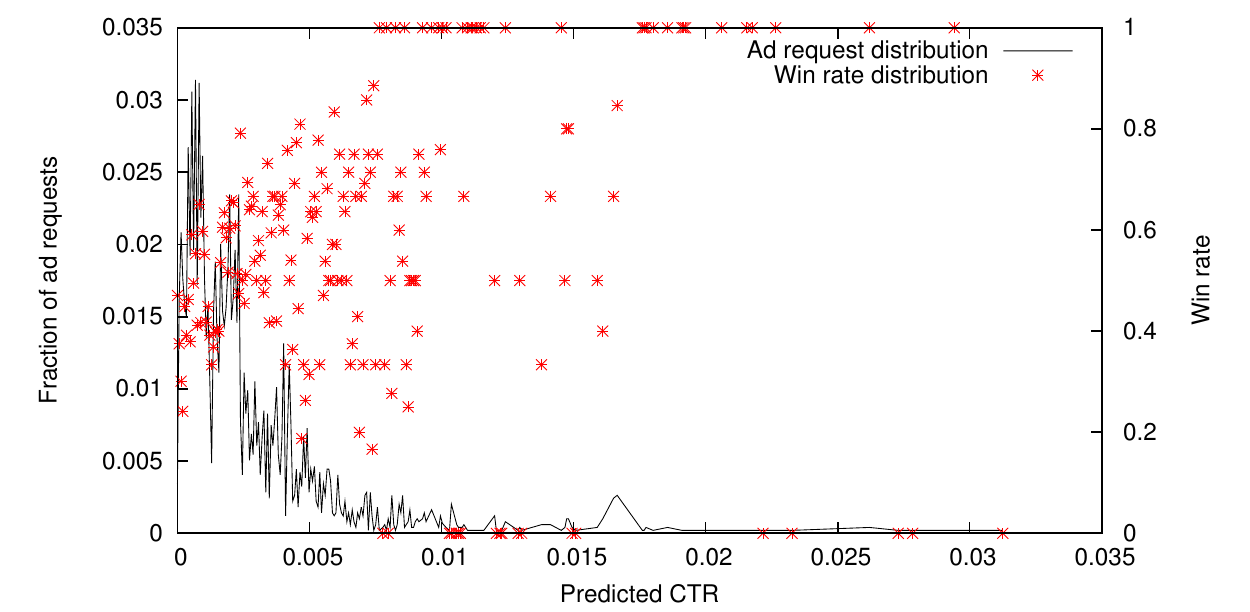}}
\caption{Offline simulation setup: simulation data is collected from real serving logs of the DSP.}
\label{exp:setup}
\end{figure*}
\begin{figure*}[htb]
\centering
\captionsetup[subfloat]{%
font=scriptsize,
labelformat=parens,labelsep=space,
listofformat=subparens}
\subfloat[Allocated budget v.s. ad request volume]{\includegraphics[width = 3.5in]{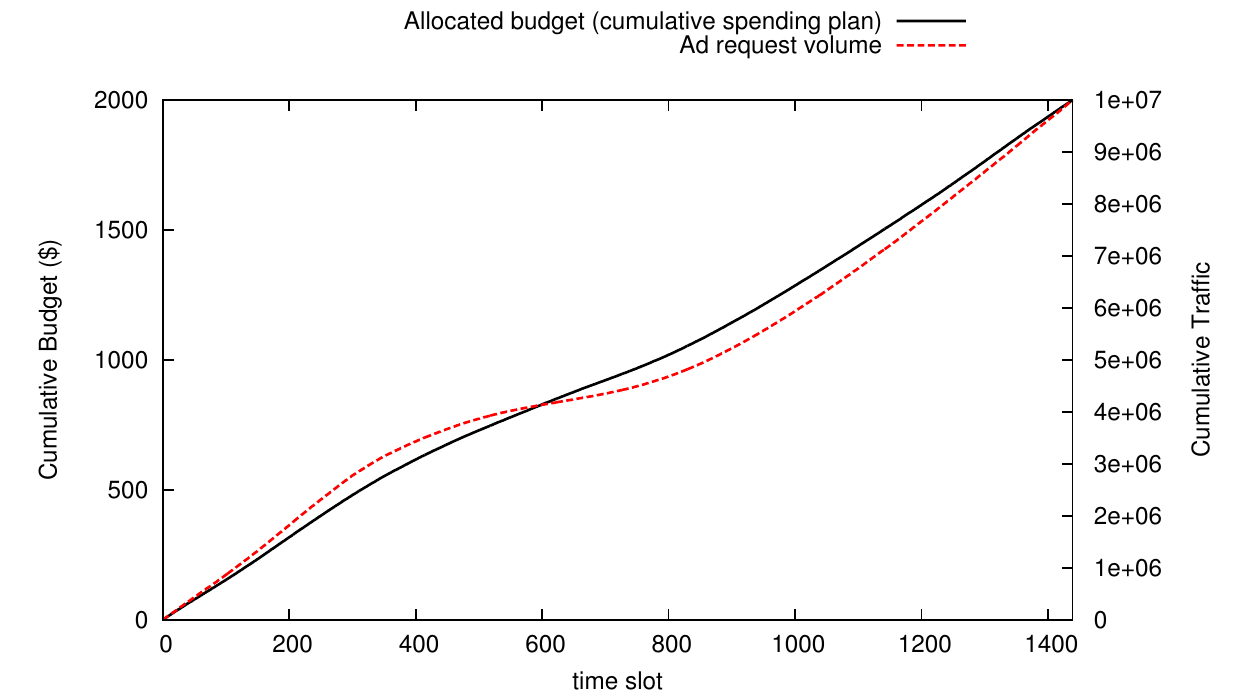}} 
\subfloat[Actual spendings and eCPCs comparison (eCPC is calculated every 15 minutes to collect sufficient number of clicks)]{\includegraphics[width = 3.5in]{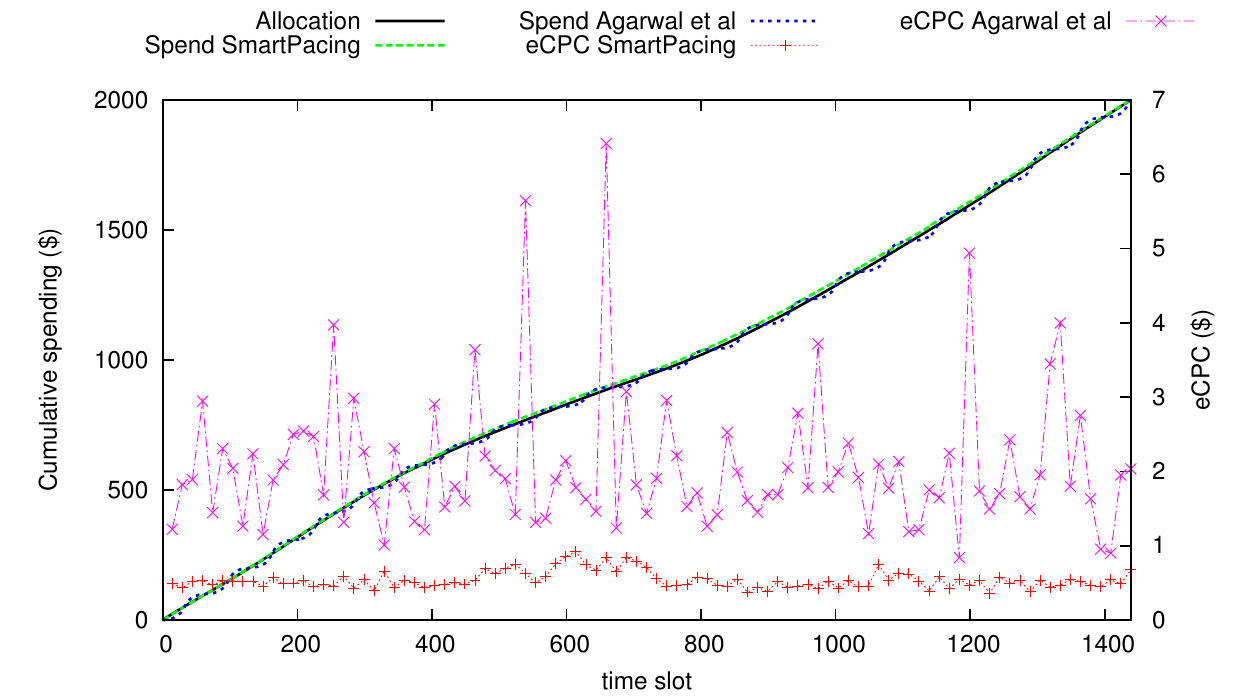}}
\caption{Comparison with state-of-the-art approach with arbitrary spending plan}
\label{exp:budgeted}
\end{figure*}

\begin{figure*}[!htb]
\centering
\captionsetup[subfloat]{%
font=scriptsize,
labelformat=parens,labelsep=space,
listofformat=subparens}
\subfloat[Spendings over time with different \# of layers (Budget:\$2,000. Budget spending plan is compromised when \# of layers is excessive.)]{\includegraphics[width = 3.5in]{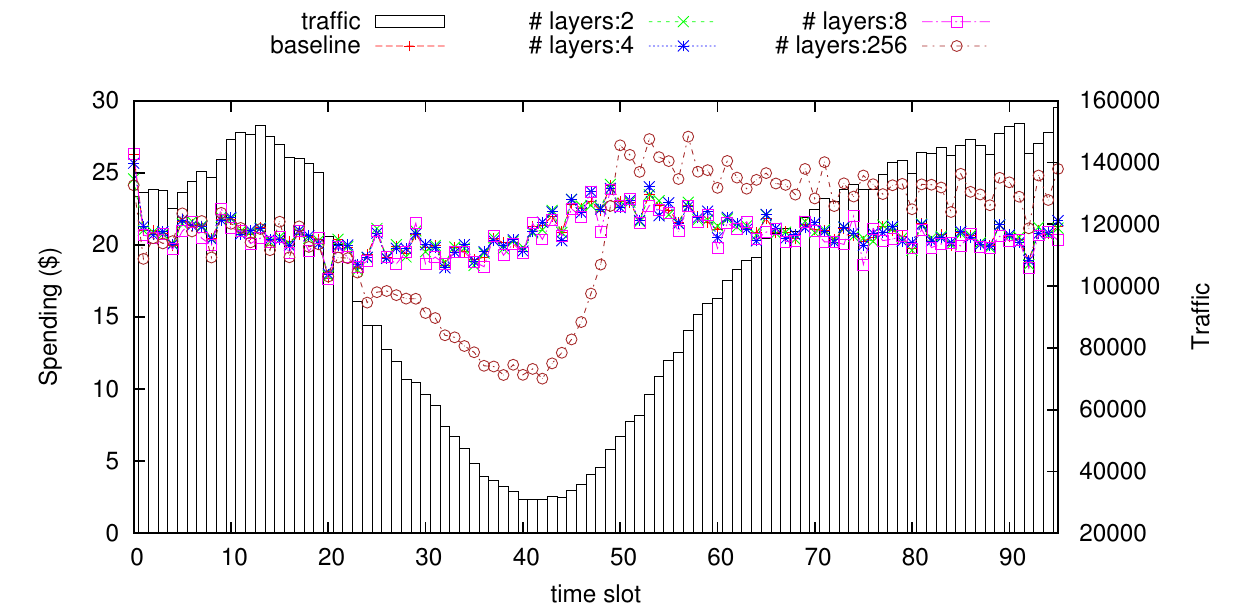}} 
\subfloat[Spendings over time with different budget (\# of layers: 8)]{\includegraphics[width = 3.5in]{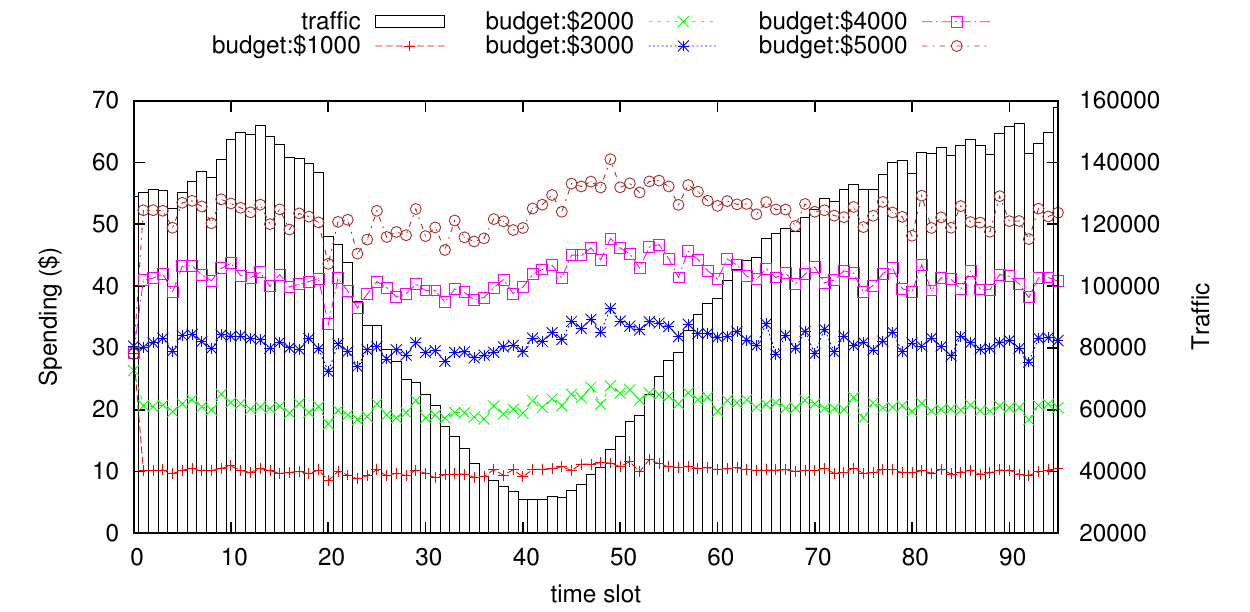}}\\
\subfloat[eCPC over time with different \# of layers (Budget:\$2,000)]{\includegraphics[width = 3.5in]{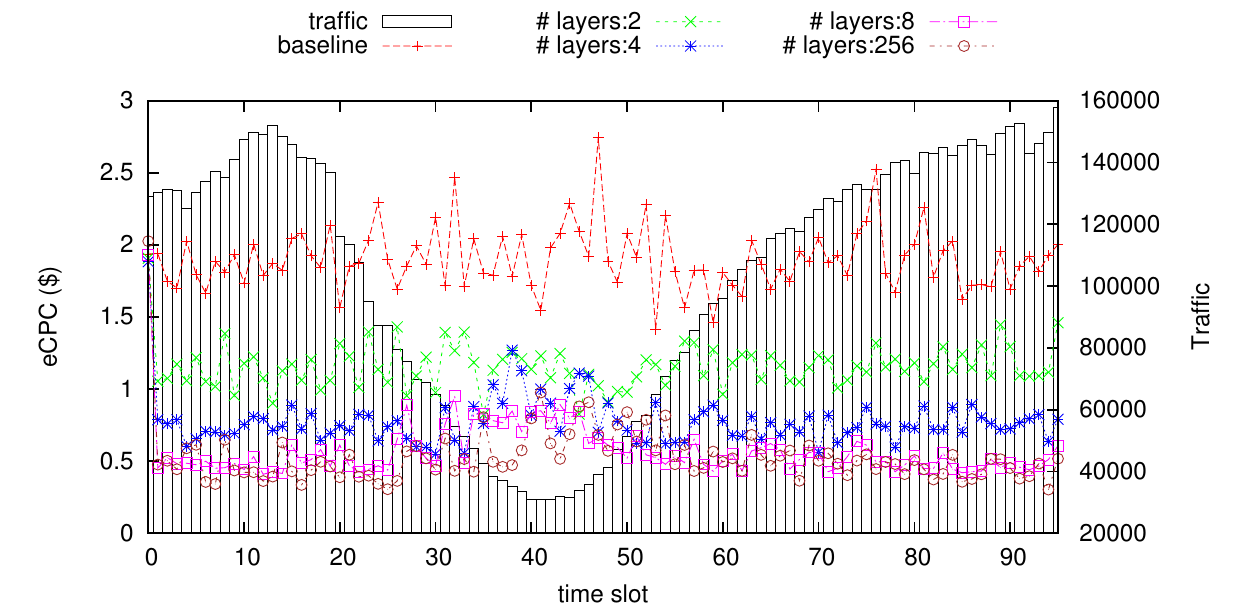}}
\subfloat[eCPC over time  with different budget (\# of layers: 8)]{\includegraphics[width = 3.5in]{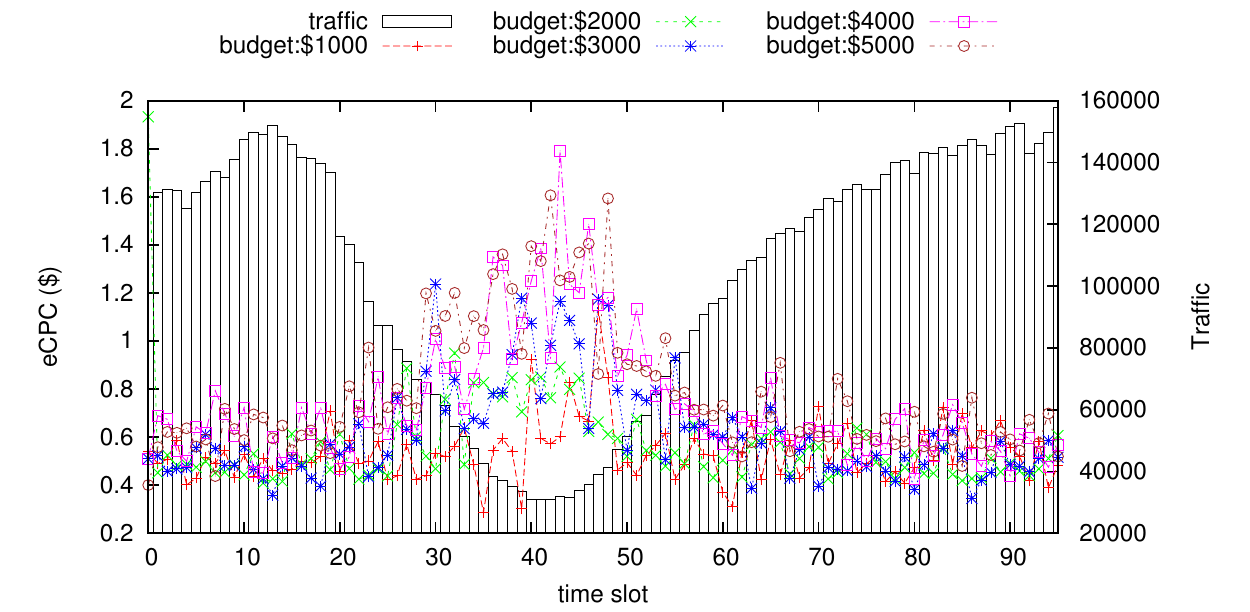}} 
\caption{Simulation result on campaign without performance goal}
\label{exp:wo_goal}
\subfloat[Spendings over time with different \# of layers (Budget:\$2,000)]{\includegraphics[width = 3.5in]{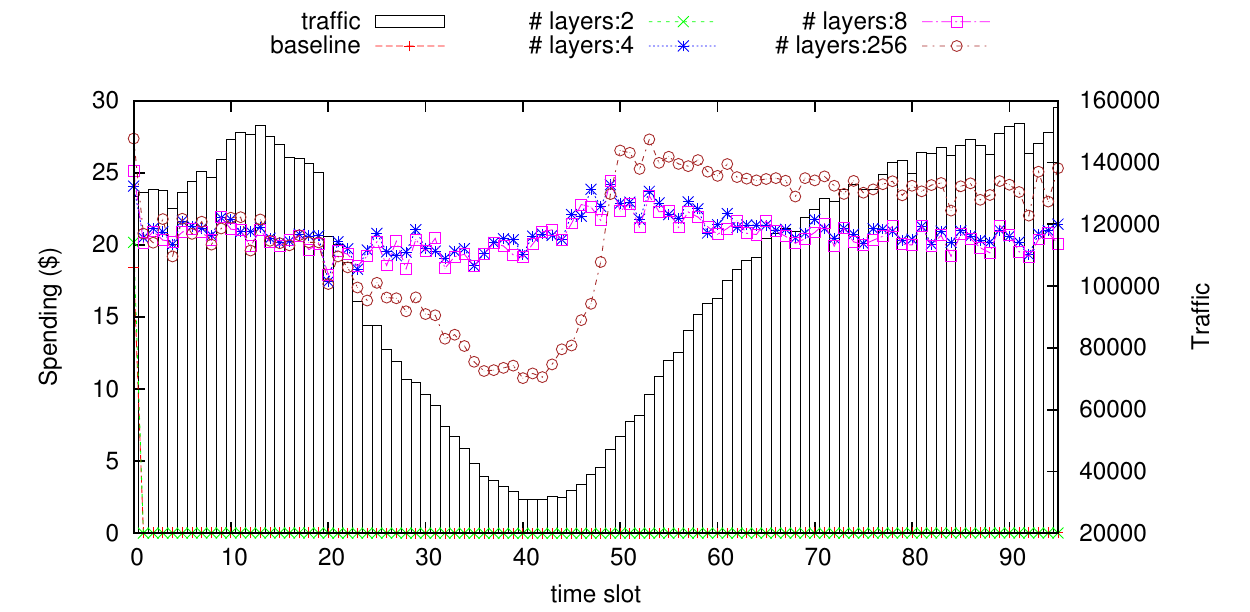}} 
\subfloat[Spendings over time with different budget (\# of layers: 8. Budget spending plan is sacrificed to meet the eCPC goal when traffic volume is low.)]{\includegraphics[width = 3.5in]{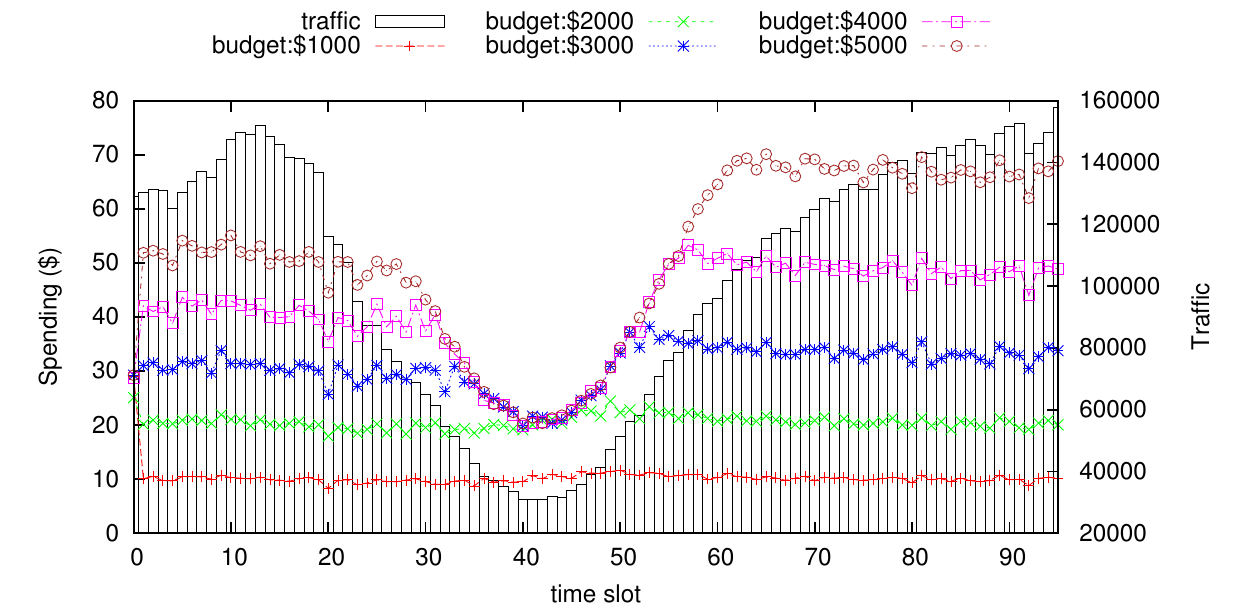}}\\
\subfloat[eCPC over time with different \# of layers (Budget:\$2,000. For baseline and \# layers=2, we plot their overall eCPC since only limited number of time slots have clicks)]{\includegraphics[width = 3.5in]{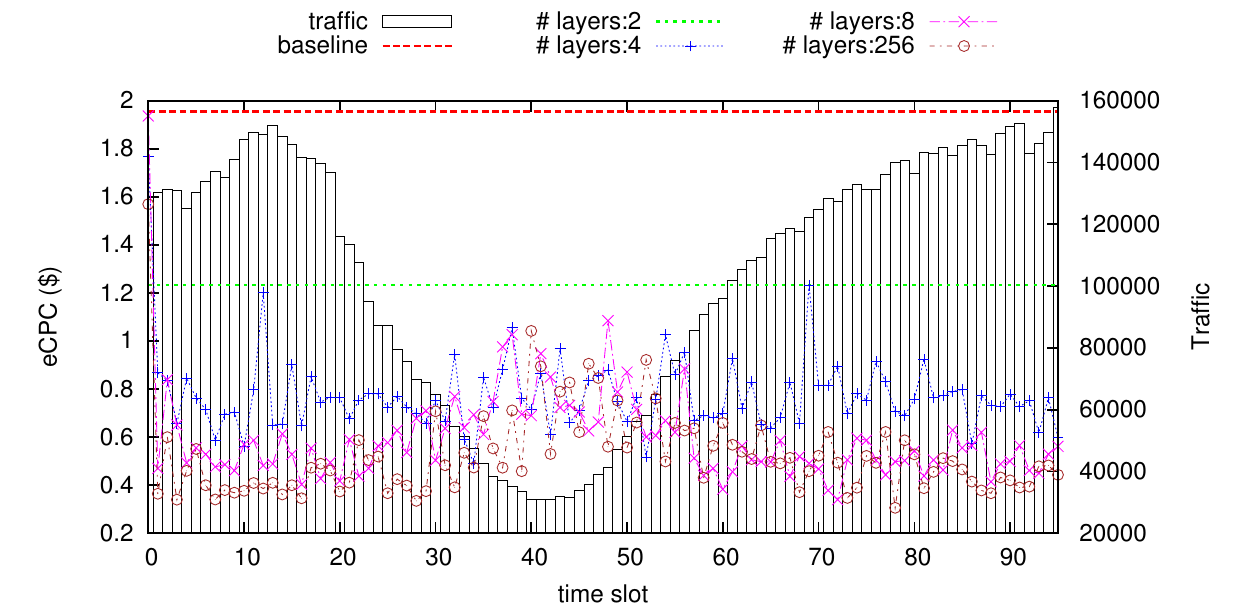}}
\subfloat[eCPC over time with different budget (\# of layers: 8)]{\includegraphics[width = 3.5in]{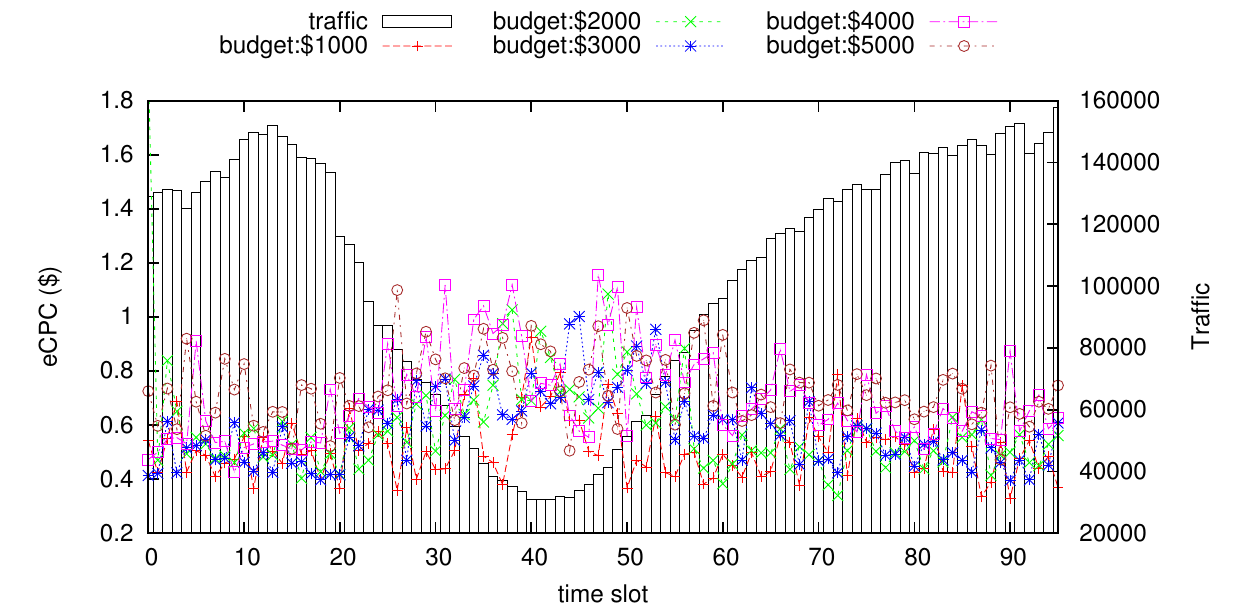}} 
\caption{Simulation result on campaign with performance goal eCPC $\le$ \$0.8}
\label{exp:wi_goal}
\end{figure*}

\subsection{Offline Simulations}
We also conduct extensive offline simulations to further assess the effectiveness of our approach. We randomly select from our demand pool an ad campaign to generate the simulation data. First, the eligible ad request distribution over 24 hours is collected (Figure \ref{exp:setup}(a)). Second, the traffic as well as the win-rate distribution over predicted CTR is collected (Figure \ref{exp:setup}(b)). We further assume the total number of ad requests is 10,000,000 and billing CPM is \$5. The campaign may or may not have a specific eCPC goal. \\
\indent We are interested in how our approach performs when compared with the state-of-the-art budget pacing method proposed by Agarwal et al \cite{agarwal2014budget}. In their method, a global pacing rate is dynamically adjusted by $\pm10\%$ every one minute to make the cumulative spending align with the allocated budget. The budget allocation is based on forecasted cumulative traffic. In our experiment, the traffic pattern in Figure \ref{exp:setup}(a) is further granulized into one minute per time slot, and we use the average traffic of previous seven days as the forecasted traffic to determine budget allocation. Figure \ref{exp:budgeted}(a) shows the cumulative traffic and allocated budget curves. Please note that the budget spending plan here is not even pacing. Since their method does not consider performance, no eCPC goal is set. We use 8 layers of pacing rates in our approach. From the result shown in Figure \ref{exp:budgeted}(b), both approaches successfully align the cumulative spending curves to the allocation curve. A subtle difference is that their spending curve has fluctuations around the allocation curve. If we look at the $AvgErr$ which captures the relative deviation in the per time slot level, their approach gets a surprisingly high error of $96\%$ while our approach generates a much smaller error of $18\%$, which means our approach can produce smoother pacing result even with very granular time slots. From the control theory perspective, their approach depends on a \emph{conventional feedback controller} while our approach essentially uses an \emph{adaptive controller} whose parameters are adaptive to the spendings in the current time slot and planned budget in the next time slot. In the RTB environment, where the \emph{plant} is complex and time-varying, an adaptive controller usually performs better than a conventional feedback controller \cite{landau2011adaptive}. Moreover, since our approach models the quality of the ad requests, it achieves a $-70\%$ lower eCPC compared to their approach. \\
\indent We continue to further study the behaviors of our method. When no eCPC goal is specified, we first fix the campaign budget at \$2,000 and vary the number of layers. As we can observe from Figure \ref{exp:wo_goal}(a), the top objective to spend out budget as much and as smooth as possible are all achieved except when the number of layers is extremely large (i.e. \# layers = 256). The reason is that an excessive number of layers delays the layer-by-layer adjustment to adapt to traffic changes. The difference of eCPCs as shown in Figure \ref{exp:wo_goal}(c) speaks of the advantage of our approach. When we increase the number of layers, we have more discernability and flexibility to \emph{cherry-pick} high performing ad requests - especially in those time slots when the ad request volume is high. Then we fix the number of layers at 8 and vary the total budget. As Figure \ref{exp:wo_goal}(b) shows, when the budget keeps going up, there are relatively less available supply so that the budget spending pattern will be more influenced by ad request volume fluctuations. Another interesting observation is that the campaign performance is better when the budget is less (Figure \ref{exp:wo_goal}(d)). This is because a small budget means our approach can cherry-pick the best performing ad requests without compromising the budget spending objectives. \\
\indent When a specific eCPC goal is specified, keeping eCPC below this goal is the most important task. Our simulation results in this scenario are shown in Figure \ref{exp:wi_goal} when the eCPC goal is set to \$0.8. Please note that whenever the pacing rate of every layer is adjusted to zero, which means the performance goal cannot be achieved based on our estimation, we reset the pacing rates of all the layers so that only the top priority layer has a $trial\ rate$. Again, we first fix budget at \$2,000 and vary the number of layers. Different from when no eCPC goal is set, we find that the spendings are extremely low when there are only 1 or 2 layers (Figure \ref{exp:wi_goal}(a)). Because in such cases, the eCPC goal can never be achieved and the pacing rates are kept being reset. It can be reconfirmed by looking at Figure \ref{exp:wi_goal}(c), in which only when number of layers is 4, 8, or 256 the average eCPC is below \$0.8. When we fix the number of layers at 8 and vary the budget, we also observe different results from when there is no eCPC goal, e.g. Figure \ref{exp:wi_goal}(b) shows quite different spending pattern from Figure \ref{exp:wo_goal}(b). The reason is that to achieve eCPC goal, sometimes we need to sacrifice budget spending objectives - especially when the traffic is low. This is even more apparent when the budget increases (Figure \ref{exp:wi_goal}(d)). We note that it is desired and is exactly one of our contributions.

\section{System Implementation}
\label{sec:implementation}
\begin{figure}
  \centering
  \includegraphics[width=3in]{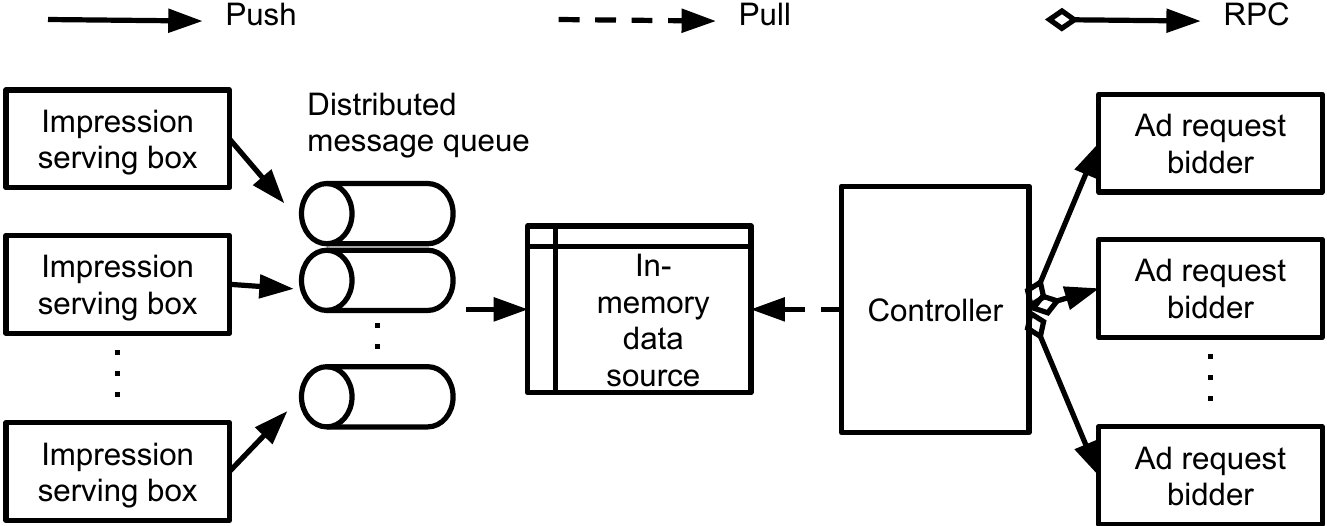}
  \caption{The implementation architecture.}
   \label{fig:system-arch}
\end{figure}

In a large scale online ad serving system, there are many infrastructure and implementation issues need to be addressed such as data consistency, service availability, and fault tolerance, that are, however, out of the scope of this paper. In this section, we mainly focus on the following challenges. \emph{rapid feedback of layered statistics:} as described in section \ref{sec:control}, the pacing rate adjustment is mainly based on the layered delivery statistics. Therefore, how to collect the online feedback data efficiently and reliably becomes a major implementation challenge. \emph{Overspending prevention:} overspending should always be avoided since it undermines either the advertiser or the DSP's interest. Thus, a \emph{quick stop} mechanism is necessary to prevent overspending. \\
\indent We address these challenges by implementing a real time feedback pipeline as well as an in-memory data source. As illustrated in Figure \ref{fig:system-arch}, impression serving boxes receive impression/click events and produce delivery messages into the message queue. The in-memory data source consumes messages from the message queue and performs aggregations on top of these messages. Finally the controller refers to the in-memory data source to send quick stop notifications or adjusted pacing rates to the bidders.

\subsection{Real Time Feedback}
Traditional ad serving systems log events such as ad requests, bids, impressions, clicks to an offline data warehouse, and perform offline processing on top of it. However, the delay is too high to fit in our scenario. Therefore, building a separate pipeline to provide real time feedback to online serving system becomes an essential requirement. \\
\indent We implement both \emph{message queue} and \emph{remote procedure call} (RPC) in our system to transfer messages. Message queue is used to send delivery messages from impression serving boxes to in-memory data source while RPC is used to send pacing rates and quick stop notifications to ad request bidders. Message queue works in asynchronous mode, which means producer and consumer are decoupled and producer has no idea whether consumer has consumed the message or not. The asynchronous nature makes it easy to achieve high throughput and low latency. RPC works in synchronous mode, which means caller will get success or failure responses from callee. Therefore, the motivation to implement both mechanisms is clear: to achieve extremely high throughput (typically billions of impressions per day), impression serving boxes do not need to track whether the messages have been consumed as long as there is no message lost. On the contrary, to make sure the pacing rates and quick stop messages are sent and applied to each bidder, the controller needs to get responses from all the bidders. \\
\indent Although we have implemented a message queue to transfer delivery messages asynchronously, the huge amount of delivery information requires some further designs. \emph{Batch processing} and \emph{micro-aggregation} are the two endeavors to further drive efficiency. At impression serving boxes side, a batch accumulates hundreds of delivery messages into a single request over the wire to reduce network overhead and amortize transmission delay. This is especially useful for high latency links such as those between data centers. At message queue side, a batch groups multiple small I/O operations into a single one to improve efficiency. Another practice we have is aggregating the delivery messages under certain conditions without information loss, e.g. aggregating impressions of the same campaign with the same predicted response rate within a certain time period.
\subsection{In-memory Data Source}
The in-memory data source stores the layered delivery information of each campaign. The main challenges of implementing such an in-memory storage are: 1) how to avoid data loss in case of failure, and 2) how to control memory usage. In order to address the potential data loss problem in system failures, we employ a snapshot plus commit log approach. We observe that the message queue itself can be a commit log as long as it honors the order of message sequence. With the help of message queue, we can easily recover the state using the message sequence number and the snapshot data. We address the memory usage issue by aggregation. As described in section \ref{sec:control}, the pacing rate adjustment is based on the delivery information of the most recent time slot and therefore only raw messages within configurable recent time need to be stored. Historical data is stored in aggregated form to minimize the memory usage.

\section{Conclusions}
We have presented a general and principled approach as well as its implementation in a real DSP system to perform smooth pacing control and maximize the campaign performance simultaneously. Experimental results showed that, compared to state-of-the-art budget pacing method, our proposed approach can significantly boost the campaign performance and achieve smooth pacing goals. \\
\indent Our future work will mainly focus on trying out different pacing schemes such as pacing based on performance with the help of supply and performance forecasting techniques, combining the proposed approach with other control methods to make it more intelligent and robust, and studying the competitions and interactions among multiple campaigns in pacing control.

\bibliographystyle{abbrv}
\bibliography{pacing.bib}

\end{document}